\documentclass{article}

\usepackage{iclr2026_conference_arxiv,times}
\iclrfinalcopy

\usepackage[utf8]{inputenc} % allow utf-8 input
\usepackage[T1]{fontenc}    % use 8-bit T1 fonts
\usepackage{hyperref}       % hyperlinks
\usepackage{url}            % simple URL typesetting
\usepackage{booktabs}       % professional-quality tables
\usepackage{amsfonts}       % blackboard math symbols
\usepackage{nicefrac}       % compact symbols for 1/2, etc.
\usepackage{microtype}      % microtypography
\usepackage{xcolor}         % colors
\usepackage[inline]{enumitem}
\usepackage{graphicx}
\usepackage{tabularx}
\usepackage{xcolor}
\usepackage{amssymb}
\usepackage{subcaption}

\newcommand{\rset}{\mathbb{R}}

\newcommand{\rmd}{\mathrm{d}}
\newcommand{\KL}{\mathrm{KL}}
\newcommand{\Id}{\mathrm{Id}}
\usepackage{amsmath,amsfonts,bm}

\def\eqref#1{equation~\ref{#1}}
\def\1{\bm{1}}

\DeclareMathAlphabet{\mathsfit}{\encodingdefault}{\sfdefault}{m}{sl}
\SetMathAlphabet{\mathsfit}{bold}{\encodingdefault}{\sfdefault}{bx}{n}

\newcommand{\vareps}{\varepsilon}

\newcommand{\calL}{\mathcal{L}}

\usepackage{tikz}
\usetikzlibrary{arrows.meta, positioning, shapes.geometric, calc}
\usepackage{comicneue} % Use the Comic Neue font 
\usepackage{upgreek}
\usepackage[most]{tcolorbox} % Load the tcolorbox package

\usepackage{algorithm}
\usepackage[noend]{algpseudocode} % For pseudocode environment, noend option removes "end if/for/while"

\usepackage[capitalise]{cleveref}
\usepackage{autonum}

\renewcommand{\eqref}[1]{\ref{#1}}
\newcommand{\MMD}{\mathrm{MMD}}

\usepackage[most]{tcolorbox}
\usepackage{lipsum} % For dummy text
\usepackage{amsmath}

\newtcbtheorem{mydef}{Definition}{
    colback=gray!15, % Background color (15% grey)
    arc=3mm,         % Radius of the rounded corners
    fonttitle=\bfseries,
    boxrule=0pt,     % No frame/border
    halign=center,   % Center the text horizontally inside the box
    sharp corners=east
}{def}

\usepackage[oxfordcolors,fancytheorems]{fancy_theorems}

\title{Learn to Guide Your Diffusion Model}

\author{%
  Alexandre Galashov  \\
  Google DeepMind \\
  Gatsby Unit, University College London \\
  \texttt{agalashov@google.com} \\
  \And
  Ashwini Pokle \\
  Google \\
  \texttt{apokle@google.com} \\
  \And
   Arnaud Doucet  \\
  Google DeepMind \\
  \texttt{arnauddoucet@google.com} \\
  \And
  Arthur Gretton  \\
  Google DeepMind \\
  Gatsby Unit, University College London \\
  \texttt{gretton@google.com} \\
  \And
  Mauricio Delbracio \\
  Google \\
  \texttt{mdelbra@google.com} \\
  \And
  Valentin De Bortoli  \\
  Google DeepMind \\
  \texttt{vdebortoli@google.com} \\
}

\begin{document}

\maketitle

\begin{abstract}

Classifier-free guidance (CFG) is a widely used technique for improving the perceptual quality of samples from conditional diffusion models. It operates by linearly combining conditional and unconditional score estimates using a \emph{guidance weight} $\omega$. While a large, static weight can markedly improve visual results, this often comes at the cost of poorer distributional alignment.
In order to better approximate the target conditional distribution,
we instead learn \emph{guidance weights} $\omega_{c,(s,t)}$, which are continuous functions of the conditioning $c$, the time $t$ from which we denoise, and the time $s$ towards which we denoise. 
We achieve this by minimizing the distributional mismatch between noised samples from the true conditional distribution and samples from the guided diffusion process.  We extend our framework to reward guided sampling, enabling the model to target distributions tilted by a reward function $R(x_0,c)$, defined on clean data and a conditioning $c$. We demonstrate the effectiveness of our methodology on low-dimensional toy examples and high-dimensional image settings, where we observe improvements in Fr\'echet inception distance (FID) for image generation. In text-to-image applications, we observe that employing a reward function given by the CLIP score leads to guidance weights that improve image-prompt alignment.
\end{abstract}

\section{Introduction}
Diffusion models \citep{sohl2015deep,ho2020denoising,song2020score,song2019generative} produce high-quality synthetic data in areas such as images \citep{saharia2022photorealistic,labs2025flux}, videos \citep{GoogleVeo2}, and proteins \citep{watson2023novo,abramson2024accurate}. These models proceed by gradually adding Gaussian noise through a diffusion process, transforming the data distribution into a Gaussian distribution. The generative model is obtained by approximating the time-reversal of this noising process. Practically, this relies on a learned denoiser network which is obtained by minimizing a regression objective. A similar strategy can, in principle, be developed to sample from a conditional distribution by learning a conditional denoiser network. 

Recently, Classifier-Free Guidance (CFG)~\citep{ho2022classifier} has become a popular choice to address the task of conditional sampling. CFG  modifies the generating process by incorporating a \emph{guidance} term, calculated as the difference between  conditional and unconditional denoiser networks. This difference is scaled by a \emph{guidance weight}, $\omega$, such that $\omega=-1$ recovers unconditional generation whereas $\omega=0$ corresponds to the conditional one. Empirically, it was shown that CFG can drastically improve performance of diffusion models compared to conditional generation  (which only uses  the conditional denoiser), especially for very high guidance weights~\citep{saharia2022photorealistic,rombach2022highresolutionimagesynthesislatent}.  \citep{chidambaram2024does,wu2024theoretical} argued that such a regime pushes the samples to the edge of support such that they become easy to classify.

Classifier-free guidance was initially justified as a method for sampling from a modified distribution $p_{\omega}(x_0|c) \propto p(x_0)^{-\omega} p(x_0|c)^{1+\omega}\propto p(x_0) p(c|x_0)^{\omega+1}$, which amplifies the conditioning term \citep{ho2022classifier}. However, subsequent work has established that the standard CFG sampling procedure does not, in fact, produce samples from this target distribution \citep{du2023reduce, koulischer2024dynamic}. Instead, CFG sampling shifts the generated samples towards the modes of the original conditional $p(x_0|c)$. While this discrepancy can be corrected with computationally intensive methods like Sequential Monte Carlo \citep{skreta2025feynman, he2025rneplugandplayframeworkdiffusion} or Markov Chain Monte Carlo \citep{du2023reduce, yazidgibbs2025, zhang2025inference}, these approaches are often impractical for large-scale applications. More importantly, naive CFG \emph{already} yields a significant quality boost and therefore such correction might not be required.

Many works, see e.g. \cite{ho2022classifier,kynkaanniemi2024applying,wang2024analysis,kim2024simple}, report improved FID scores when using some level of guidance ($\omega >  0$). This observation is at odds with the original motivation of CFG. Given that the learned denoiser $\hat{x}_\theta(x_t, c)$ is only an approximation of the true denoiser $\mathbb{E}[x_0|x_t,c]$, this suggests that CFG acts as a \emph{correction to this approximation} which leads to a better modeling of the target conditional distribution $p(x_0|c)$.

Based on this observation, we propose to \emph{learn} a guidance weight $\omega$ in order to better approximate the conditional distribution $p(x_0|c)$. We generalize the CFG method and allow the guidance weight to be both conditioning and time dependent, i.e., $\omega_{c, (s,t)}$. This weight is used to guide the denoising process from time $t$ to time $s$. We learn $\omega_{c, (s,t)}$ by matching the distribution of samples from the true diffusion process and the distribution of samples of the guided one.
Empirically, we show that our method allows us to learn $\omega_{c, (s,t)}$ that better approximates the target distribution compared to the unguided model. For image generation, this leads to a consistently lower FID~\citep{heusel2017gans} than the unguided model or a model with a constant guidance weight.

 We further develop our method in a setting where an additional reward function $R(x_0,c)$ is defined on samples from the diffusion model and a conditioning $c$, and we want to bias the samples from the model to the regions of high rewards $R(x_0,c)$.
 In text-to-image applications, we found that empowering our method with a reward function given by the CLIP score~\citep{clip_score}, leads to guidance weights which improve both FID and image-prompt alignment.

The paper is organized as follows. In~\Cref{sec:background}, we provide a short background on the necessary concepts. In~\Cref{sec:learning_to_guide}, we describe our approach of learning guidance weights $\omega_{s,(s,t)}$. In~\Cref{sec:related_work}, we present the related work on the topic. Finally, the experimental results are presented in~\Cref{sec:experimental_results}.

\section{Background}
\label{sec:background}

\paragraph{Notation.} We write $p_t$ for a probability distribution $p(x_t)$, and $p_{s|t}$ to denote a conditional distribution $p(x_s |x_t)$ for any $s,t$. We also use $p_{0|c}$ to denote $p(x_0|c)$, $p_{s|t,0}$ for $p(x_s|x_t,x_0)$, and $p_{s|t,c}$ for $p(x_s|x_t,c)$. We denote by $p_{0,c}$ a joint distribution $p(x_0,c)$. Finally, we write $\mathcal{N}(x;\mu,\Sigma)$ to denote the Gaussian density of argument $x$, mean $\mu$ and covariance $\Sigma$ and $\mathcal{N}(\mu,\Sigma)$ for the distribution.

\paragraph{Diffusion models.} The goal of conditional diffusion models is to sample from a target conditional distribution $p_{0|c}$ on $\rset^d$, where $c \sim p(c)$ is a conditioning signal (i.e. a text prompt) sampled from a conditioning distribution. We adopt here the Denoising Diffusion Implicit Models (DDIM) framework of \citet{song2020denoising}. Let $x_{t_0}\sim p_{0|c}$  and define for $0 = t_0 < \dots < t_N = 1$ the process $x_{t_1:t_N}:=(x_{t_1},...,x_{t_N})$ by
\begin{equation}\label{eq:forward}
    \textstyle
   p(x_{t_1:t_N}|x_{t_0})= p_{t_N|t_0}(x_{t_N}|x_{t_0}) \prod_{k=1}^{N-1} p_{t_{k}|t_{k+1},0}(x_{t_{k}} | x_{t_{k+1}}, x_{t_0}),
\end{equation}
where, for $0\leq s<t\leq 1$, we have
\begin{equation}
    \textstyle
    \label{eq:transition_density}
    p_{s|t,0}(x_{s} |x_{t},x_0)=\mathcal{N}(x_{s} ; \mu_{s,t}(x_0, x_{t}), \Sigma_{s,t}).
\end{equation}
Both the mean $\mu_{s,t}(x_0, x_t)$ and covariance $\Sigma_{s,t}$ depend on a \emph{churn} parameter $\varepsilon \in [0,1]$, which controls the amount of stochasticity. Their full expressions are given in (\eqref{eq_app:mean_sigma}) in~\Cref{app_sec:ddim}.
Here, the mean and variance are selected so that for any $t \in [0,1]$, we recover the \emph{noising process}
\begin{equation}
\label{eq:interpolation}
\textstyle
p_{t|0}(x_t | x_0) = \mathcal{N}(x_t ; \alpha_t x_0, \sigma_t^2 \Id) ,
\end{equation}
for $\alpha_t$ and $\sigma_t$ such that $\alpha_0=1, \sigma_0 =0$ and $\alpha_1 = 0,\sigma_1 = 1$, i.e. $p_{1|0}(x_1|x_0) = \mathcal{N}(x_1 ; 0,\Id)$. More precisely for any $0 \leq s \leq t \leq 1$, we have 
\begin{equation}
\label{eq:first_consistency}
\textstyle
    p_{s|0}(x_s|x_0) = \int p_{t|0}(x_t|x_0)p_{s|t,0}(x_s|x_t,x_0) \rmd x_t . 
\end{equation}
\paragraph{Sampling with DDIM.}
At inference time, $x_{t_0}\sim p_{0|c}$ is generated by starting from a Gaussian $x_{t_N}\sim \mathcal{N}(0,\Id)$ and 
$x_{t_k}\sim p(\cdot|x_{t_{k+1}},c)$ for $k=N-1,...,0$, where for $0\leq s<t\leq 1$
\begin{equation}
    \textstyle
    \label{eq:backward_kernel}
    p_{s|t,c}(x_s|x_t,c) = \int p_{s|t,0}(x_s|x_t,x_0) p_{0|t,c}(x_0 |x_t,c) \rmd x_0,
\end{equation}
where $p_{0|t,c}(x_0 |x_t,c)$ is a posterior distribution, see also \citep{song2020denoising} for discussion.
Since we do not have access to $p_{0|t,c}$, we approximate it by $\updelta_{\hat{x}_{\theta}(x_t,c)}$ for any $t \in [0, 1]$, where $\hat{x}_{\theta}(x_t,c)\approx \mathbb{E}[x_{0}|x_t,c]$
is a neural network denoiser with parameters $\theta$ trained by minimizing the loss
\begin{equation}
    \textstyle
    \label{eq:loss_function}
        \mathcal{L}(\theta) = \int_0^1 \lambda(t) \mathbb{E}_{(x_0,c)\sim p_{0,c}, x_t \sim p_{t|0}}[\| x_0 - \hat{x}_\theta(x_t, c) \|^2] \rmd t.
\end{equation}
Here, $\lambda(t) \geq 0$ is a weighting function (see \citep{kingma2021variational}). This yields the following procedure to sample approximately from $p_{0|c}$, by starting from $x_{t_N} \sim \mathcal{N}(0,\Id)$, we then follow
\begin{align}
    \textstyle
    \label{eq:sampling_procedure}
    x_{t_k} \sim \mathcal{N}(\mu_{t_{k},t_{k+1}}(\hat{x}_{\theta}(x_{t_{k+1}}, c),x_{t_{k+1}}),\Sigma_{t_{k},t_{k+1}}), \qquad k=N-1,...,0
\end{align}
\paragraph{Classifier-Free Guidance (CFG).} Let $\hat{x}_{\theta}(x_t, \varnothing)\approx \mathbb{E}[x_0|x_t]$ be an unconditional denoiser learned using a regression objective~(\eqref{eq:loss_function}) (where conditioning is omitted). 
CFG replaces $\hat{x}_{\theta}(x_t, c)$ when simulating approximately from $p_{0|c}$ by
\begin{equation}
    \textstyle
    \label{eq:cfg_delta_term}
    \hat{x}_{\theta}(x_t, c; \omega)=\hat{x}_{\theta}(x_t, c) +\omega   \Delta_{\theta}(x_t,c),
\end{equation}
where $\omega$ is the \emph{guidance weight} and $\Delta_{\theta}(x_t,c) = \hat{x}_{\theta}(x_t, c) -  \hat{x}_{\theta}(x_t, \varnothing)$.
In order to sample the data with CFG, we first sample $x_{t_N} \sim \mathcal{N}(0,\Id)$ and then follow
\begin{align}
    \textstyle
    \label{eq:sampling_procedure_guided}
    x_{t_k} \sim \mathcal{N}(\mu_{t_{k},t_{k+1}}(\hat{x}_{\theta}(x_{t_{k+1}}, c; \omega),x_{t_{k+1}}),\Sigma_{t_{k},t_{k+1}}),\quad  k=N-1,...,0.
\end{align}
When $\omega = 0$, the sampling procedure~(\eqref{eq:sampling_procedure_guided}) is equivalent to \emph{conditional} sampling~(\eqref{eq:sampling_procedure}), while $\omega=-1$ corresponds to a procedure using only the unconditional denoiser, allowing to sample approximately from $p_0$. In practice, one sets $\omega>0$ in order to emphasize conditioning.

In this paper, we generalize CFG and make \emph{guidance weight} $\omega_{c, (s, t)}$ a function of conditioning $c$ and of two time-steps $s <t$, where $t$ is the \emph{proposal} timestep from which we denoise towards a \emph{target} timestep $s$ as in~(\eqref{eq:backward_kernel}). Denoising following~(\eqref{eq:sampling_procedure_guided}) corresponds to setting, $s=t_{k}$ and $t=t_{k+1}$. We denote $\boldsymbol{\omega}=(\omega_{c,(s,t)})_{s,t\in[0,1],t>s}$. In~\Cref{alg:sampling_diffusion_ddim_guided}, we describe the CFG sampling method with DDIM.

\paragraph{Maximum Mean Discrepancy.} To optimize the guidance weights $\boldsymbol{\omega}$, we  match the true diffusion distribution $p$ and the guided $p^{\boldsymbol{\omega}}$, see~\Cref{sec:learning_to_guide}. We employ the Maximum Mean Discrepancy (MMD)~\citep{gretton2012kernel} with energy kernel \citep{SzeRiz04,sejdinovic2013equivalence},
\begin{equation}
    \textstyle
\MMD_{(\beta,\lambda)}[p^{\boldsymbol{\omega}},p]= \mathbb{E}_{p_{\boldsymbol{\omega}} \otimes p}[||x-y||^\beta_2] - \frac{\lambda}{2} (\mathbb{E}_{p^{\boldsymbol{\omega}} \otimes p^{\boldsymbol{\omega}}} [||x-x'||^\beta_2] + \mathbb{E}_{p\otimes p}  [||y-y'||^\beta_2] ),
\label{eq:generalized_mmd}
\end{equation}
for independent $x,x' \sim p^{\boldsymbol{\omega}}$ and $y,y'\sim p$, with $\beta \in (0,2)$ and $\lambda \in [0,1].$ We require $\lambda=1$ for a valid MMD, however $\lambda<1$ may be preferable for generative modeling \citep{bouchacourt2016disco,debortoli2025distributional}; other characteristic kernels may also be used \citep{sriperumbudur2010hilbert}.

\section{Learning to guide your diffusion model}
\label{sec:learning_to_guide}

In this section, we introduce our method for learning the guidance weights, $\boldsymbol{\omega}$, assuming access to  a pre-trained conditional denoiser, $\hat{x}_{\theta}( x_t, c)$, and an unconditional one, $\hat{x}_{\theta}(x_t, \varnothing)$. Our approach is based on enforcing \emph{consistency conditions}--- that any valid diffusion process must satisfy.
We first derive a theoretically sound objective from \emph{marginal consistency} condition, which is however impractical to optimize due to high variance. To overcome this, we introduce a stronger, more practical condition, which we call \emph{self-consistency}. Enforcing \emph{self-consistency} is sufficient for achieving \emph{marginal consistency} and, crucially, results in a simple, low-variance objective. We formulate our complete approach for learning guidance weights based on this simplified approach. We present alternative approaches in~\Cref{app_sec:other_approaches}.

\subsection{Consistency conditions}

\paragraph{Marginal consistency.} The process defined by~(\eqref{eq:forward}) admits the marginal distribution for $s \in (0, 1]$
\begin{equation}
    \textstyle
    \label{eq:forward_inteprolatnt}
    p_{s}(x_s) = \int p_{s|0}(x_s | x_0) p_{0,c}(x_0, c) \rmd x_0 \rmd c.
\end{equation}
This marginal distribution can also be obtained for any $0\leq s<t\leq1$ via
\begin{equation}
    \textstyle
    \label{eq:true_s_t_marginal}
    p_{s}(x_s) = \int\int \int p_{s|t,c}(x_s|x_t, c) p_{t|0}(x_t|x_0) p_{0,c}(x_0, c) \rmd x_t  \rmd x_0 \rmd c,
\end{equation}
with $p_{s|t,c}$ given by DDIM~(\eqref{eq:backward_kernel}). The equality (\eqref{eq:true_s_t_marginal}) states that the marginal at time $s$ can be obtained by sampling $(x_0,c)\sim p_{0,c}$, noising $x_0$ to $x_t|x_0\sim p_{t|0}$ and then denoising it back to time $s$ with $p_{s|t,c}$. 

We now consider a denoising mechanism relying on the guided denoiser approximation~(\eqref{eq:cfg_delta_term}) which follows the construction in (\eqref{eq:true_s_t_marginal}) to obtain a sample at time $s$. Again we sample $(x_0,c)\sim p_{0,c}$ and $x_t \sim p_{t|0}(\cdot|x_0)$. However, the denoising to time $s$ where $0\leq s<t\leq1$ is done with the guided denoiser approximation, so that the marginal distribution of the resulting sample at time $s$ is
\begin{equation}
    \textstyle
    \label{eq:guided_backwards}
    p_{s}^{t,(\theta,\boldsymbol{\omega})}(x_s) = \int\int \int p_{s|t,c}^{(\theta,\boldsymbol{\omega})}(x_s|x_t, c) p_{t|0}(x_t|x_0) p_{0,c}(x_0, c) \rmd x_t  \rmd x_0 \rmd c,
\end{equation}
where
\begin{equation}
    \textstyle
    \label{eq:guided_backwards_transition}
    p_{s|t,c}^{(\theta,\boldsymbol{\omega})}(x_s|x_t, c) = p_{s|t,0}(x_s|x_t, \hat{x}_{\theta}(x_t, c; \boldsymbol{\omega})).
\end{equation}
The distribution $p_{s}^{t,(\theta,\boldsymbol{\omega})}$~(\eqref{eq:guided_backwards}) is typically not equal to $p_s$~(\eqref{eq:forward_inteprolatnt}) as we used a delta-function approximation~(\eqref{eq:guided_backwards_transition}) with a model $\hat{x}_{\theta}(x_t, c; \boldsymbol{\omega})$ instead of sampling from $p_{0|t,c}$ as required by (\eqref{eq:backward_kernel}). Nevertheless, we could  attempt to find guidance weights $\boldsymbol{\omega}$ to satisfy \emph{marginal consistency}, i.e. for all $0 \leq s < t \leq 1$
\begin{equation}
    \textstyle
    \label{eq:marginal_consistency}
   p_{s}^{t,(\theta,\boldsymbol{\omega})}(x_s)  \approx p_{s}(x_s).
\end{equation}
This could be achieved by minimizing the MMD~(\eqref{eq:generalized_mmd})
\begin{equation}
    \textstyle
    \label{eq:marginal_objective}
    \mathcal{L}_m(\boldsymbol{\omega}) =  \mathbb{E}_{(s, t) \sim p(s,t)} [
    \MMD_{(\beta,\lambda)}[p_s^{t, (\theta,\boldsymbol{\omega})}(\cdot),p_s(\cdot)] ],
\end{equation}
where $p(s,t)$ is a distribution on $0\leq s<t\leq 1$. This distribution is crucial for good empirical performance, and we discuss it in detail in~\Cref{sec:guidance_method}.
The gradient of~(\eqref{eq:marginal_objective}) could suffer from high variance due to marginalization over $(x_0,c)$, however, which could make it challenging to optimize. We found that this approach did not work in practice. Below, we propose a simpler, lower variance method,  albeit one that imposes more constraints on the guided backward scheme compared to~(\eqref{eq:marginal_objective}).
\vspace{-0.1in}
\paragraph{Self-consistency (conditioning on $(c,x_0)$).}
For a fixed $(x_0,c) \sim p(x_0,c)$, we denote by
\begin{equation}
    \textstyle
    \label{eq:guided_conditional}
    p_{s|0,c}^{t,(\theta,\boldsymbol{\omega})}(x_s | x_0, c) = \int p_{s|t,c}^{(\theta,\boldsymbol{\omega})}(x_s|x_t, c) p_{t|0}(x_t|x_0) \rmd x_t,
\end{equation}
which is the term under integral in~(\eqref{eq:guided_backwards}) depending on $(x_0,c)$. We consider a \emph{self-consistency} condition
\begin{equation}
    \textstyle
    \label{eq:guidance_self_consistency}
    p_{s|0,c}^{t,(\theta,\boldsymbol{\omega})}(x_s | x_0, c) \approx 
    p_{s|0,c}(x_s|x_0,c)=p_{s|0}(x_s|x_0),
\end{equation}
where $p_{s|0,c}^{t,(\theta,\boldsymbol{\omega})}(\cdot | x_0, c)$ is given by~(\eqref{eq:guided_conditional}) and $p_{s|0}(\cdot|x_0)$ is a \emph{noising process}~(\eqref{eq:interpolation}). We used a notation $p_{s|0,c}(x_s|x_0,c)=p_{s|0}(x_s|x_0)$ to highlight that $x_0 \sim p_{0|c}(\cdot|c)$. Intuitively, this condition means that as we start from $x_0 | c$ and go through the \emph{noising process} $x_t | x_0$ and then denoise with guidance to time $s < t$, the distribution at time $s$ should be the same as of the \emph{noising process}.

The condition~(\eqref{eq:guidance_self_consistency}) is much stronger than~(\eqref{eq:marginal_consistency}). Indeed, if~(\eqref{eq:guidance_self_consistency}) is satisfied for every $(x_0,c)\sim p_{0,c}$, then by integrating it over $(x_0,c)$, we will get~(\eqref{eq:marginal_consistency}). The reverse is not true.
Moreover, the lack of dependence on $x_t$ (and on $x_0)$ in guidance weights $\omega_{c,(s,t)}$ makes it very unlikely for this condition to hold. However, it provides a learning signal for guidance weights and we will aim to satisfy it approximately.

For $(x_0,c)$ in the support of $p(x_0,c)$, we could approximately satisfy~(\eqref{eq:guidance_self_consistency}) by minimizing wrt $\boldsymbol{\omega}$, $\MMD_{(\beta,\lambda)}[p_{s|0,c}^{t,(\theta,\boldsymbol{\omega})}(\cdot | x_0,c), p_{s|0,c}(\cdot|x_0)]$, see~(\eqref{eq:generalized_mmd}).
Averaging over all $(x_0,c)\sim p(x_0,c)$ leads to
\begin{equation}
    \textstyle
    \label{eq:guidance_loss_function}
    \mathcal{L}_{\beta,\lambda}(\boldsymbol{\omega}) = \mathbb{E}_{(x_0, c) \sim p_{0,c}, s,t \sim p(s,t)} [ \MMD_{(\beta,\lambda)}[p_{s|0,c}^{t,(\theta,\boldsymbol{\omega})}(\cdot | x_0,c), p_{s|0,c}(\cdot|x_0)] ].
\end{equation}
This approach does not suffer from high variance compared to~(\eqref{eq:marginal_objective}),
since both $c$ and $x_0$ are fixed. Furthermore, we found that using~(\eqref{eq:guidance_loss_function}) works well in practice (see our experiments in~\Cref{sec:experimental_results}).

\subsection{Learning to guide}
\label{sec:guidance_method}
We present here our approach for learning guidance weights $\boldsymbol{\omega}$ based on the \emph{self-consistency} loss~(\eqref{eq:guidance_loss_function}). We also provide empirical evaluation of other approaches in~\Cref{sec:experimental_results}.

\vspace{-0.1in}
\paragraph{Guidance learning objective.} 
For $(x_0,c) \sim p_{0,c}(x_0,c)$, we sample $x_s \sim p_{s|0}(\cdot|x_0)$, i.e., $x_s \sim \mathcal{N}(\alpha_s x_0,\sigma^2_s \Id)$.
We also sample $\tilde{x}_s(\boldsymbol{\omega}) \sim p_{s|0,c}^{t,(\theta,\boldsymbol{\omega})}$, i.e., sample $\tilde{x}_t \sim \mathcal{N}(\alpha_t x_0,\sigma^2_t \Id)$ then 
\begin{align}
    \label{eq:guided_forward_sample_2}
    &\tilde{x}_s(\boldsymbol{\omega}) \sim \mathcal{N}(\mu_{s,t}(\hat{x}_{\theta}(\tilde{x}_t, c; \omega_{c,(s,t)}), \tilde{x}_t), \Sigma_{s,t}),
\end{align}
where $\mu_{s,t}$ and $\Sigma_{s,t}$ are given by DDIM (see~(\eqref{eq_app:mean_sigma}) in~\Cref{app_sec:ddim}). 
The objective~(\eqref{eq:guidance_loss_function}) can be written as
\begin{equation}
    \textstyle
    \label{eq:distributional_objective}
    \mathcal{L}_{\beta,\lambda}(\boldsymbol{\omega}) = \mathbb{E}_{(x_0, c) \sim p_{0,c}, s,t \sim p(s,t)} [
    \mathbb{E}[||\tilde{x}_s(\boldsymbol{\omega}) - x_s||^\beta_2] - \frac{\lambda}{2} \mathbb{E} [ ||\tilde{x}_s(\boldsymbol{\omega}) - \tilde{x}'_s(\boldsymbol{\omega})||^\beta_2]
    ],
\end{equation}
where we dropped terms not depending on $\boldsymbol{\omega}$, and with independent  $\tilde{x}_s(\boldsymbol{\omega}),\tilde{x}'_s(\boldsymbol{\omega}) \sim p_{s|0,c}^{t,(\theta,\boldsymbol{\omega})}$.

\paragraph{Simplified guidance learning objective.} A special case of~(\eqref{eq:distributional_objective}) with $\beta=2$ and $\lambda=0$, leads to
\begin{equation}
    \textstyle
    \label{eq:l2_objective}
    \mathcal{L}_{\ell_2}(\boldsymbol{\omega}) = \mathcal{L}_{\beta,0}(\boldsymbol{\omega})= \mathbb{E}_{(x_0, c) \sim p_{0,c}, s,t \sim p(s,t)} \left[
    \mathbb{E}[||\tilde{x}_s(\boldsymbol{\omega}) - x_s||^2_2]
    \right].
\end{equation}
We found that~(\eqref{eq:l2_objective}) was very effective but more sensitive to hyperparameters than~(\eqref{eq:distributional_objective}). This approach is however cheaper than~(\eqref{eq:distributional_objective}) since it avoids quadratic complexity $O(m^2)$ of computing interaction terms. We refer the reader to~\Cref{app_sec:l2_approach} and to~\Cref{app_alg:l2_algorithm} for more details on the use of~(\eqref{eq:l2_objective}).

\vspace{-0.1in}
\paragraph{Guidance network.} The guidance weights $\omega_{c,(s,t)}^\phi=\omega(s,t, c;\phi)$ are given by a neural network with parameters $\phi$ . We use ReLU activation at the end to prevent negative guidance weights. For more details, see~\Cref{app_sec:experimental_details}. We denote $\boldsymbol{\omega}^\phi=(\omega^\phi_{c,(s,t)})_{s,t\in[0,1],t>s}$ and employ $\mathcal{L}_{\beta,\lambda}(\phi)$ instead of $\mathcal{L}_{\beta,\lambda}(\boldsymbol{\omega})$.

\vspace{-0.1in}
\paragraph{Distribution $p(s,t)$.}
We choose target time $s$ to be distributed as $s \sim \mathcal{U}[S_{\min},1-\zeta- \delta]$. We define $\Delta t \sim \mathcal{U}[\delta,  1-\zeta-s]$ and we let $t=s+\Delta t$.
Here, $S_{\min}$ controls the minimal time, and $\zeta$ controls how close it gets to $1$. 
The parameter $\delta$ controls the distance between time-steps and we found it to be very important, see~\Cref{fig:imagenet_ablation}. Even though during inference $|t-s|\approx \frac{1}{T}$ is typically small ($T$ is a number of steps), we found that using larger $\delta \approx 0.1$ during training worked better in practice.

\vspace{-0.1in}
\paragraph{Empirical objective.} We sample $\{x^i_0,c^i\}_{i=1}^n \overset{\mathrm{i.i.d.}}{\sim} p_{0,c}$ from the training set and we additionally sample noise levels $\{s_i\}_{i=1}^{n} \overset{\mathrm{i.i.d.}}{\sim} \mathcal{U}[S_{\min},1-\zeta- \delta]$, as well as time increments $\{\Delta t_{i}\}_{i=1}^n \overset{\mathrm{i.i.d.}}{\sim} \mathcal{U}[\delta, 1-\zeta -s_i]$ and we let $t_i = s_i + \Delta t_{i}$. For each $(x^i_0, s_i)$, we sample $m$ ``particles'' $\{x_{s_{i}}^{j} \}_{j=1}^m \overset{\mathrm{i.i.d.}}{\sim} p_{s_i | 0}(\cdot | x_0^i)$ from the \emph{noising process}~(\eqref{eq:interpolation}), which defines the target samples. We  also produce $m$ ``particles'' $\{\tilde{x}_{s_{i}}^{j}(\boldsymbol{\omega}^\phi)\}_{j=1}^m \overset{\mathrm{i.i.d.}}{\sim} p_{s_i|0, c_i}^{t_i, (\theta,\boldsymbol{\omega}^\phi)}(\cdot|x_0^i, c_i)$ by first sampling $\{\tilde{x}^j_{t_{i}}\}_{j=1}^m \overset{\mathrm{i.i.d.}}{\sim}  p_{t_{i}| 0}(\cdot | x_0^i)$ from the \emph{noising process}~(\eqref{eq:interpolation}) and then denoising with guidance and DDIM using~(\eqref{eq:guided_forward_sample_2}). This defines the proposal samples.
We expand loss function~(\eqref{eq:distributional_objective}) as a function of guidance network parameters $\phi$ defined on the empirical batches as follows (where  $\lambda\in [0,1]$ and $\beta\in(0,2)$, see~\Cref{alg:learning_to_guide})
\begin{equation}
    \textstyle
    \label{eq:guidance_empirical_loss}
    \hat{\mathcal{L}}_{\beta,\lambda}(\phi) = \frac{1}{n}\sum_{i=1}^{n}\left[ \frac{1}{m}\sum_{j=1}^{m} ||\tilde{x}_{s_{i}}^{j}(\boldsymbol{\omega}^\phi) - x_{s_{i}}^{j}||_{2}^\beta  -\frac{\lambda}{2} \frac{1}{m(m-1)}\sum_{j\neq k} ||\tilde{x}_{s_{i}}^{j}(\boldsymbol{\omega}^\phi) -\tilde{x}_{s_{i}}^{k}(\boldsymbol{\omega}^\phi)||_{2}^\beta \right]
\end{equation}

\begin{algorithm}
\caption{Learning to Guide}
\begin{algorithmic}[1] % The [1] enables line numbers
    \State \textbf{Input:} Init. guidance parameters $\phi$; (frozen) denoiser $\hat{x}_\theta$; data distribution $p_0$; learning rate $\eta$; $\zeta > 0$, $S_{\min}>0$, $\delta>0$, b.s. $n$, n. of particles $m$, $\lambda \in [0,1]$,$\beta \in [0,2]$, DDIM churn $\varepsilon \in [0,1]$.
    \Repeat
        \State Sample batch of clean data and their conditionings $\{x^i_0,c^i\}_{i=1}^n \overset{\mathrm{i.i.d.}}{\sim} p_{0,c}$.
        \State Sample $\{s_i\}_{i=1}^{n} \overset{\mathrm{i.i.d.}}{\sim} \mathcal{U}[S_{\min},1-\zeta- \delta]$, $\{\Delta t_{i} \}_{i=1}^n \overset{\mathrm{i.i.d.}}{\sim} \mathcal{U}[\delta, 1-\zeta-s_i]$, let $t_i=s_i + \Delta t_i$
        \State (\textbf{True process}) Sample $m$ particles $\{x_{s_{i}}^{j}\}_{j=1}^m  \overset{\mathrm{i.i.d.}}{\sim} p_{s_i | 0}(\cdot | x_0^i)$ from noising process~(\eqref{eq:interpolation})
        \State (\textbf{Guided process}) Sample $m$ particles $\{\tilde{x}^j_{t_{i}}\}_{j=1}^m \sim p_{t_{i}| 0}(\cdot | x_0^i)$ from noising process~(\eqref{eq:interpolation})
        \State Compute guidance weights $\omega_{i} = \omega^\phi_{c_i,(s_i,t_i)}$
        and $\hat{x}_{\theta}(\tilde{x}^j_{t_i}, c_i; \omega_i)$ using~(\eqref{eq:cfg_delta_term})
        \State Sample $\tilde{x}_{s_i}^j(\boldsymbol{\omega}^\phi)\sim p_{s_i|t_i,0}(\cdot |\tilde{x}^j_{t_{i}}, \hat{x}_{\theta}(\tilde{x}^j_{t_i}, c_i; \omega_i))$ from DDIM~(\eqref{eq:transition_density}) with \emph{churn} parameter $\varepsilon$
        \State (\textbf{Loss}) Compute loss
        \State $\hat{\mathcal{L}}_{\beta,\lambda}(\phi) = \frac{1}{n}\sum_{i=1}^{n}\left[ \frac{1}{m}\sum_{j=1}^{m} ||\tilde{x}_{s_{i}}^{j}(\boldsymbol{\omega}^\phi) - x_{s_{i}}^{j}||_{2}^\beta  -\frac{\lambda}{2} \frac{1}{m(m-1)}\sum_{j\neq k} ||\tilde{x}_{s_{i}}^{j}(\boldsymbol{\omega}^\phi) -\tilde{x}_{s_{i}}^{k}(\boldsymbol{\omega}^\phi)||_{2}^\beta \right]$
        \State Update $\phi \leftarrow \phi - \eta \nabla_{\phi} \hat{\mathcal{L}}_{\beta,\lambda}(\phi)$
    \Until{convergence}
    \State \textbf{Output:} Optimized guidance network parameters $\phi$
\end{algorithmic}
\label{alg:learning_to_guide}
\end{algorithm}
\vspace{-0.2in}

\paragraph{Learning to guide with rewards.} CFG can be used to produce samples with high reward $R(x_0,c)$, set by a practitioner. Denoising with guidance from $t$ to $s$ as described by~(\eqref{eq:guided_forward_sample_2}), gives us an approximation $\hat{x}_{\theta}(x_t, c; \omega^\phi_{c,(s,t)})$ of clean data. 
We could use it to optimize guidance weights, by defining the loss $\calL_{R}(\phi) = -\mathbb{E}_{(s,t)\sim p(s,t), x_0,c \sim p(x_0, c), x_t \sim p_{t|0}(\cdot|x_0)}\left[R \left(\hat{x}_{\theta}(x_t, c; \omega^\phi_{c,(s,t)}), c \right) \right]$.
Directly minimizing this loss may lead to \emph{reward hacking}~\citep{reward_hacking}. Thus, we regularize this objective using $\hat{\mathcal{L}}_{\beta,\lambda}$. For reward weight $\gamma_{R} \geq 0$, we optimize $\calL_{\text{tot}}(\phi) =  \hat{\mathcal{L}}_{\beta,\lambda}(\phi) + \gamma_{R} \calL_{R}(\phi)$ (see~\cref{alg:learning_to_guide_with_rewards}).

\section{Related work}
\label{sec:related_work}

\paragraph{Classifier-Free Guidance.} Using guidance in the sampling of diffusion models was first investigated by~\cite{dhariwal2021diffusion}. They proposed a classifier guidance method which linearly combines the unconditional score estimate and the input gradient of the log-probability of a (time-varying) classifier. To avoid training such classifier on noisy training data, \citet{ho2022classifier} proposed Classifier-Free Guidance (CFG), which linearly combines a conditional and unconditional denoisers. By varying the guidance weight, one is able to obtain high-quality samples. This approach has become prominent in the literature, see~\citet{adaloglou2024cfg} for a recent introduction.

CFG has been extended in many different directions, some of them proposing dynamic mixing strategies to control the guidance weight throughout the sampling process (e.g. \citet{sadat2023cads,kynkaanniemi2024applying,wang2024analysis,shen2024rethinking,malarz2025classifier,li2024self,koulischer2024dynamic,koulischer2025feedback,xia2025rectified, sadat2024eliminating, zheng2024characteristic}). Another line of research focuses on removing the need for the unconditional score component by using either an ``inferior" version of the conditional model to provide a negative guidance signal \citep{karras2024guiding,adaloglou2025guidingdiffusionmodelusing}, by leveraging time-step information~\citep{sadat2024no}, or by enabling the model to act as its own implicit classifier \citep{tang2025diffusion}. Other approaches focus on fixing some of the limitations of classifier-free guidance; e.g., \citet{chung2024cfg++} introduced CFG++, a modification to the standard CFG to address off-manifold issues, while \citet{koulischer2025feedback} show that the true conditional distribution can be obtained by approximating a term corresponding to the derivative of a R\'enyi divergence. 
Very recently, \citet{fan2025cfg} introduced CFG-Zero$\ast$, a CFG variant for flow-matching models. It leverages an optimized scale factor and a zero-init technique (skipping initial ODE steps) to correct for early velocity inaccuracies, resulting in improved text-to-image/video generation.
\emph{We highlight that our approach could be seamlessly combined with any other guidance technique by simply replacing the definition of the guided denoiser~(\eqref{eq:cfg_delta_term}).}
Finally, there exists a large literature on correcting CFG with Sequential Monte Carlo~\citep{skreta2025feynman, he2025rneplugandplayframeworkdiffusion} or MCMC techniques \citep{du2023reduce, yazidgibbs2025, zhang2025inference}. These approaches are mostly orthogonal to other improvements of CFG, and can be combined with most of the dynamic mixing strategies (i.e.~\citet{malarz2025classifier}). 

CFG has also been reinterpreted as a \textit{predictor-corrector} in \cite{bradley2024classifier} and analyzed theoretically in a variety of works \citep{fu2024unveil,wu2024theoretical,chidambaram2024does,kong2024diffusion,frans2025diffusion}. \cite{pavasovic2025understanding} have shown that while CFG can overshoot the target distribution in low dimension, it can reproduce the target distribution in high dimensions.

Concurrently to our work, \citet{papalampidi2025dynamicclassifierfreediffusionguidance} introduced dynamic classifier guidance by selecting the guidance weight among a list of pre-determined guidance weights using online evaluators.

\vspace{-0.1in}
\paragraph{Learning CFG weight.} \citet{azangulov2025adaptivediffusionguidancestochastic} introduced an algorithm based on stochastic control to optimize the guidance weights. However, this algorithm is not scalable as it requires guided backwards trajectories to estimate the gradient.
Concurrently to us, \citet{yehezkel2025navigating}  introduced an annealing guidance scheduler that dynamically adjusts the guidance weight based on the timestep  and the magnitude of the conditional noise discrepancy. The method learns an adaptive policy to better balance image quality and alignment with the text prompt throughout the generation process. One of their losses is similar to an alternative approach we explored -- guided score matching (see~\cref{app_sec:guided_score_matching}). We found that optimizing such loss led to guidance weights equal to zero.

\vspace{-0.1in}
\paragraph{Distillation.} Our method shares similarities with distillation approaches such as Diff-Instruct~\citep{diff_instruct}, Variational Score Distillation~\citep{vsd_paper} and Moment Matching Distillation (MMD)~\citep{salimans2024multistep}, which distill a diffusion model by approximately minimizing the KL divergence between the distilled generator and the pretrained teacher model. One could think about our method as a form of distillation where the KL divergence is replaced by a scoring rule, the pretrained teacher model is replaced by the true distribution, and the student is replaced by a guided diffusion. Our methodology could be extended to the distillation setting.

\vspace{-0.1in}
\paragraph{Distributional approaches for diffusion models.} Our objective function was motivated by~\citep{debortoli2025distributional}, where (\eqref{eq:generalized_mmd}) is used to learn $p(x_0|x_t,c)$ to obtain distributional diffusion models. Our method is also related to Inductive Moment Matching (IMM)~\citep{zhou2025inductivemomentmatching}, where a generative model is trained by enforcing marginal consistency~(\eqref{eq:marginal_consistency}).
Instead of directly optimizing for~(\eqref{eq:marginal_consistency}), their objective minimizes the distance between $p_{s}^{t,(\theta_n,\boldsymbol{\omega})}$ and $p_{s}^{r,(\theta_{n-1},\boldsymbol{\omega})}$, where $n$ is an iteration number and $s<r<t$ is an intermediate time between $s$ and $t$. However, they do not marginalize over $(x_0,c)\sim p_{0,c}$ as in~(\eqref{eq:guided_backwards}); instead, they sample a batch of $(x_0,c)$ and use it for both $p_{s}^{t,(\theta_n,\boldsymbol{\omega})}$ and $p_{s}^{r,(\theta_{n-1},\boldsymbol{\omega})}$. This means they optimize an objective similar to self-consistency~(\eqref{eq:guidance_self_consistency}), where they use a batch of $(x_0,c)$ instead of single points.

\section{Experimental results}
\label{sec:experimental_results}

In this section, we present experimental results on learning guidance weights. In~\Cref{sec_exp:imagenet_results_distribution}, we provide results for image generation benchmarks -- ImageNet $64 \times 64$~\citep{imagenet_paper} and CelebA~\citep{liu2015faceattributes} with resolution $64 \times 64$. We provide results on text-to-image (T2I) benchmark - MS-COCO 2014 \citep{lin2014microsoft} ($10$K images) at $512 \times 512$ resolution in~\Cref{sec_exp:text_to_image_results}. Finally, due to space constraints, we provide 2D Mixture of Gaussians (MoG) results in~\Cref{sec_exp:2d_results}.

\subsection{Image generation}
\label{sec_exp:imagenet_results_distribution}

\paragraph{Experimental setting.} We evaluate the performance of our method on ImageNet $64 \times 64$ and CelebA with resolution $64 \times 64$. As evaluation metrics, we use FID~\citep{heusel2017gans} and Inception Score (IS)~\citep{salimans2016improved}. First, we pretrain diffusion models on ImageNet and CelebA. We then freeze these models and train guidance network $\omega^\phi_{c,(s,t)}$ via~\Cref{alg:learning_to_guide}. We also train a variant of our method where conditioning is omitted, i.e. $\omega^\phi_{(s,t)}$.
On top of that, we also train the simplified $\ell_2$ method using~\Cref{app_alg:l2_algorithm} using the same methodology. We report metrics based on $50k$ samples. Please refer to~\Cref{app_sec:experimental_details} for more details.

\paragraph{Baselines.} We report performance of the unguided model as well as the model with a constant guidance. We further report performance of limited interval guidance (LIG)~\citep{kynkaanniemi2024applying}. Guidance scale and intervals were selected via grid search for the lowest FID (see~\Cref{app_sec:experimental_details}).

\begin{table}[h!]
\centering
\caption{\textbf{ImageNet 64x64}. We report FID and IS for different methods (the best are in \textbf{bold}).}
\vspace{-0.1in}
\label{tab:imagenet_results}
\begin{tabular}{lccc}
\toprule
\textbf{Method Name} & \textbf{Guidance Weight} & \textbf{FID $\downarrow$} & \textbf{Inception Score (IS) $\uparrow$} \\
\midrule
& \textbf{Baselines} \\
\midrule
Unguided & $\omega = 0$ & 4.46 & 43.52 \\
Constant guidance & $\omega = 0.25$ & 2.40 & 66.72 \\
Limited interval guidance & $\omega(t) = 0.95$ for $t \in [0.2,0.8]$ & \textbf{2.11} & \textbf{71.60} \\
\midrule
& \textbf{Learned guidance approaches} \\
\midrule
Self-consistency~(\eqref{eq:distributional_objective}) & $\omega^\phi_{c,(s,t)}$ & \textbf{1.99} & $73.62$ \\
Self-consistency~(\eqref{eq:distributional_objective}) & $\omega^\phi_{(s,t)}$ & $2.07$ & $76.7$ \\
$\ell_2$ objective~(\eqref{eq:l2_objective})  & $\omega^\phi_{c,(s,t)}$ & 2.09 & 75.93 \\
$\ell_2$ objective~(\eqref{eq:l2_objective}) & $\omega^\phi_{(s,t)}$ & $2.10$ & \textbf{77.55} \\
\bottomrule
\end{tabular}
\end{table}
\vspace{-0.1in}

\paragraph{Results.} The results for ImageNet $64 \times 64$ are given in~\Cref{tab:imagenet_results} and for CelebA  in~\Cref{tab:celeba_results}. In both cases, the LIG baseline outperforms the other baselines in terms of FID and Inception Score. Our self-consistency approach~(\eqref{eq:distributional_objective}) leads to the best results, especially when the conditioning information is provided. This highlights the importance of adjusting guidance weights for different conditioning.

\begin{figure}[h]
    \centering
    \includegraphics[width=.9\linewidth]{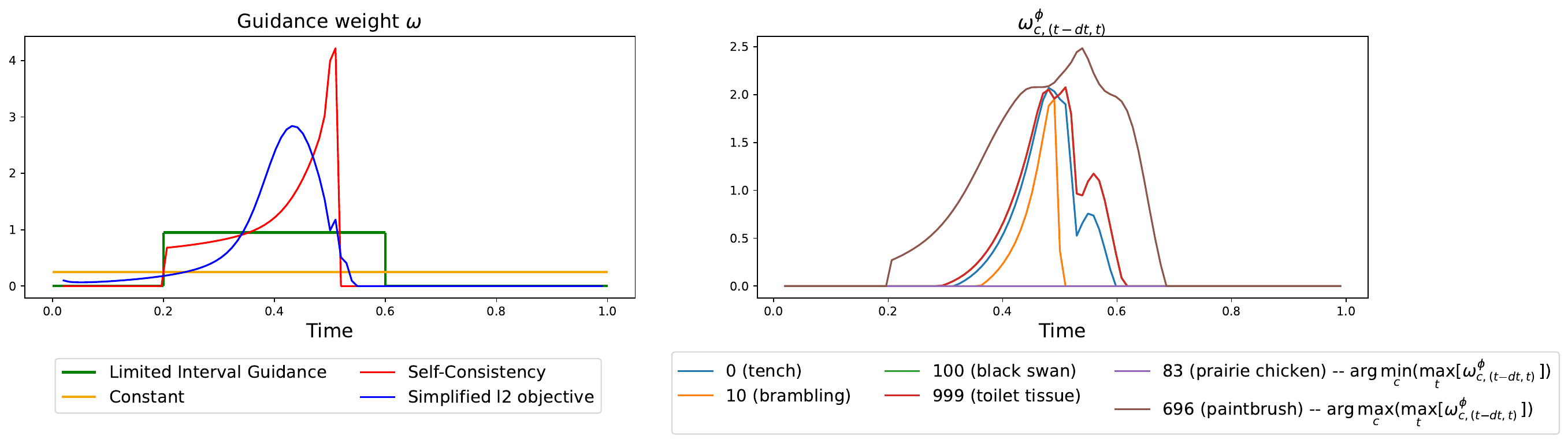}
    % \vspace{-0.1in}
    \caption{\textbf{Learned guidance weights on ImageNet 64x64}. \textbf{Left}, guidance weights  $\omega^\phi_{(t-dt,t)}$ (conditioning-agnostic) for baselines as well as for self-consistency~(\eqref{eq:distributional_objective}) and $\ell_2$~(\eqref{eq:l2_objective}) objectives, where $dt=1/100$. X-axis is time. \textbf{Right}, guidance weights $\omega^\phi_{c,(t-dt,t)}$ for specific ImageNet classes.
    }
    \label{fig:guidance_weights_imagenet}
    % \vspace{-0.2in}
\end{figure}

In~\Cref{fig:guidance_weights_imagenet}, left, we visualize the learned conditioning-agnostic guidance weights $\omega^\phi_{(t-dt,t)}$ with $dt=1/100$, for self-consistency~(\eqref{eq:distributional_objective}) and $\ell_2$~(\eqref{eq:l2_objective}) approaches, as well as for the baselines. Learned guidance weights seem to be positive on a similar interval as LIG, but the shape of the weights is quite different. In~\Cref{fig:guidance_weights_imagenet}, right, we visualize guidance weights $\omega^\phi_{c,(t-dt,t)}$ for different classes of ImageNet, learned by self-consistency~(\eqref{eq:distributional_objective}). First, we arbitrary choose some classes --  0 (tench), 10 (brambling), 100 (black swan) and 999 (toilet tissue). Then, we visualize the ones which achieve the largest and the lowest guidance weights, i.e.
83 (prairie chicken) = $\arg\min_{c} [\max_t \omega^\phi_{c,(t-dt,t)}]$ and 696 (paintbrush) = $\arg\max_{c} [\max_t \omega^\phi_{c,(t-dt,t)}]$. For the class $83$ (prairie chicken), the guidance weight is zero. For the class $696$ (paintbrush) it has quite a different behavior compared to others, being more aggressive and positive on a larger interval. Overall, this, variability highlights importance of adjusting guidance weights per conditioning.

\begin{table}[h!]
\centering
\caption{\textbf{CelebA $64 \times 64$}. We report FID and IS for different methods (the best are in \textbf{bold}).}
\vspace{-0.1in}
\label{tab:celeba_results}
\begin{tabular}{lccc}
\toprule
\textbf{Method Name} & \textbf{Guidance Weight} & \textbf{FID $\downarrow$} & \textbf{Inception Score (IS) $\uparrow$} \\
\midrule
& \textbf{Baselines} \\
\midrule
Unguided & $\omega = 0$ & 2.44 & 2.94 \\
Constant guidance & $\omega = 0.01$ & 2.45 & 2.94 \\
Limited interval guidance & $\omega(t) = 0.7$ for $t \in [0.0,0.8]$ & \textbf{2.37} & \textbf{2.96} \\
\midrule
& \textbf{Learned guidance approaches} \\
\midrule
Self-consistency~(\eqref{eq:distributional_objective}) & $w^\phi_{c,(s,t)}$ & $\textbf{2.10}$ & $\textbf{2.98}$ \\
Self-consistency~(\eqref{eq:distributional_objective}) & $w^\phi_{(s,t)}$ & $2.28$ & $2.97$ \\
$\ell_2$ objective~(\eqref{eq:l2_objective}) & $w^\phi_{c,(s,t)}$ & 2.36 & 2.95 \\
$\ell_2$ objective~(\eqref{eq:l2_objective}) & $w^\phi_{(s,t)}$ & $2.33$ & $2.95$ \\
\bottomrule
\end{tabular}
\end{table}
\vspace{-0.15in}

\begin{figure}[h]
    \centering
    \includegraphics[width=.9\linewidth]{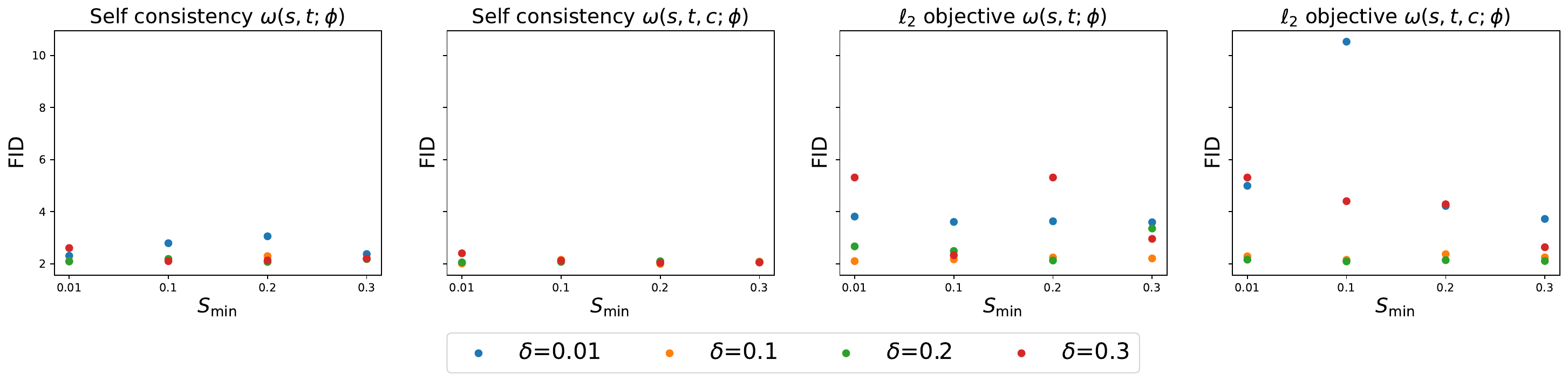}
    % \vspace{-0.1in}
    \caption{\textbf{Ablation over $\delta$ and $S_{\min}$ on ImageNet $64 \times 64$.} On the X-axis we report values of $S_{\min}$ and on Y-axis we show FID. 
    Each column denotes a method while a color corresponds to a value of $\delta$.
    }
    \label{fig:imagenet_ablation}
    % \vspace{-0.2in}
\end{figure}

\paragraph{Ablations.} In~\Cref{fig:imagenet_ablation}, we study impact of  $\delta$ and $S_{\min}$ on ImageNet $64 \times 64$. We always use $\zeta=0.01$ which introduces a small "safety margin" over the original diffusion model (i.e. $t \in [\zeta,1-\zeta]$ instead of $t \in[0,1]$). The self-consistency approach~(\eqref{eq:guidance_self_consistency}) is less sensitive to the parameters compared to $\ell_2$~(\eqref{eq:l2_objective}). Overall, very small $\delta=0.01$ leads to worse performance compared to larger ones, motivating us to train with large gaps ($\delta \approx0.1$) between $s$ and $t$. This finding is surprising because during sampling $|s-t| \sim 0.01$ for $100$ sampling steps. We hypothesize that a larger gap $|s-t|$ provides a more stable and informative gradient signal for the guidance network, which then successfully generalizes to the small-step intervals used during inference thanks to the smoothness of the network $\omega^\phi_{c,(\cdot,\cdot)}$.
The performance is not very sensitive to $S_{\min}$, but we found $S_{\min}=0.2$ worked the best.

\subsection{Text-to-image generation}
\label{sec_exp:text_to_image_results}
We evaluate our method on MS COCO 2014 dataset at $512\times512$ resolution for text-to-image (T2I) task. We pretrain a 1.05B parameter flow matching model~\citep{lipman2022flow} that uses a Multimodal Diffusion Transformer backbone~\citep{labs2025flux}.
We freeze it
and train a guidance network 
$\omega^\phi_{c_{\text{CLIP}}, c_{\text{T5}, (s,t)}}$
via~\Cref{alg:learning_to_guide}, where $c_{\text{CLIP}}$ and $c_{\text{T5}}$ denote CLIP~\citep{radford2021learning} and T5~\citep{raffel2020exploring} embeddings of the text prompt. As baseline, we consider manually selected guidance weight $\omega$.
Instead of an empty prompt $\varnothing$ for the
unconditional term~(\eqref{eq:cfg_delta_term}), we replace it with a fixed negative prompt $c_{\text{neg}} =$ ``blurred, blurry, disfigured, ugly, tiling, poorly drawn''. This 'negative guidance' setup is for both our learned model and 
the baseline.
We also consider a setting with a reward function $R(x_0,c)$ given by CLIP score~\citep{clip_score} computed between image $x_0$ and prompt $c$. We
train guidance network via~\Cref{alg:learning_to_guide_with_rewards} with $\gamma_R=10^5$.
We report FID and CLIP Score using $10$K samples. For more  experimental details, see~\Cref{app_sec:experimental_details}.
\begin{table}[h!]
\centering
\caption{\textbf{MS COCO $512 \times 512$}. FID and CLIP score for different methods (the best are in \textbf{bold}).}
\vspace{-0.1in}
\begin{tabular}{lccc}
\toprule
\textbf{Method Name} & \textbf{Guidance Weight} & \textbf{FID $\downarrow$} & \textbf{CLIP Score $\uparrow$} \\
\midrule
& \textbf{Baselines} \\
\midrule
Unguided & $\omega = 0$ & \textbf{24.74} & 0.278 \\
Constant guidance & $\omega = 7.5$ & 31.2 & \textbf{0.306} \\
\midrule
& \textbf{Our approaches} \\
Self-consistency~(\eqref{eq:distributional_objective}) & $\omega^\phi_{c_{\text{CLIP}}, c_{\text{T5}, (s,t)}}$ & \textbf{18.01} & $0.295$ \\
Self-consistency~(\eqref{eq:distributional_objective}) + CLIP Score reward & $\omega^\phi_{c_{\text{CLIP}}, c_{\text{T5}, (s,t)}}$ & 28.37 & \textbf{0.306} \\
\bottomrule
\end{tabular}
\label{table:ms-coco-results}
\end{table}

\begin{figure}[h]
    \centering
    \begin{minipage}[c]{0.4\textwidth}
    \includegraphics[width=\textwidth]{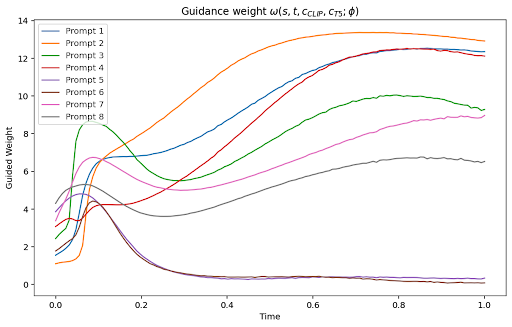}
    \end{minipage}\hfill
    \begin{minipage}[c]{0.54\textwidth}
    \scriptsize 
    \textbf{Prompt 1}. Man performing stunt on a skateboard on a road.\\
    \textbf{Prompt 2}. A street with cars and construction workers working.\\
    \textbf{Prompt 3}. A bird is sitting on a branch among unfocused trees.\\
    \textbf{Prompt 4}. A wooden table topped with four white bowls.\\
    \textbf{Prompt 5}. A light brown horse's face is shown at close range. \\
    \textbf{Prompt 6}. A living room with furniture, a fireplace, and a large scenic window.\\
    \textbf{Prompt 7}. The large bird has a red face and black feathers.\\
    \textbf{Prompt 8}. A small dog on TV behind the words: "What did I do wrong?"
    \end{minipage}
    \vspace{-0.1in}
    \caption{\textbf{Learned guidance weights on MS COCO $512\times512$} trained with self-consistency~(\eqref{eq:distributional_objective}) and CLIP reward loss.
    Please refer to \Cref{fig:ms-coco-visualization,fig:ms-coco-visualization-additional} for the corresponding images.
    }
    \label{fig:guidance_weights_coco}
\end{figure}

\vspace{-0.15in}
\paragraph{Results.} The quantitative results are summarized in~\Cref{table:ms-coco-results}.
Our method outperforms unguided and guided baselines in terms of FID, which is consistent to image experiments. However, it achieves a lower CLIP score than a guided baseline. Adding CLIP score reward leads a similar CLIP score as guided baseline, but achieves lower FID.
We provide qualitative results in Figures~(\ref{fig:ms-coco-visualization})-(\ref{fig:ms-coco-visualization-additional_2}), see~\Cref{app_sec:t2i_samples}. Our method generates images that are more realistic and better aligned with the text prompts. The learned guidance weights are shown in~\Cref{fig:guidance_weights_coco}. We observe high variability depending on the prompt.

\vspace{-0.1in}

\section{Discussion}

In this paper, we presented an approach to learn CFG weights $\omega_{c,(s,t)}$ as a function of conditioning $c$ and times $s$ and $t$, using the self-consistency condition~(\eqref{eq:guidance_self_consistency}). This rather strong condition is motivated by a weaker marginal consistency condition~(\eqref{eq:marginal_consistency}), which is satisfied by the true backwards diffusion process. Our approach yields guidance weights that improve FID on image generation tasks -- ImageNet $64\times64$ and CelebA $64\times64$. Our analysis reveals that guidance weights vary significantly depending on the conditioning information, implying that CFG with learnable, conditioning-dependent weights can improve conditional sampling performance.

We extended our methodology to text-to-image tasks with a reward function given by the CLIP score. We found that our approach leads to highly variable prompt-dependent guidance weights and visually provides better prompt alignment compared to baselines. Quantitatively, however, we found that the performance was close to a baseline with a manually selected guidance weight function.

Future work will focus on theoretical understanding of our objective function and its guidance solutions. Moreover, we will explore alternative reward functions for better prompt-image alignment, and investigate the impact of different guidance approaches. We hope our work motivates further research into how time- and conditioning-dependent guidance weights affect the sampled distributions.

\bibliographystyle{apalike}
\bibliography{references}

%%%%%%%%%%%%%%%%%%%%%%%%%%%%%%%%%%%%%%%%%%%%%%%%%%%%%%%%%%%%
\newpage
\appendix

\section*{Organization of the appendix}

In~\Cref{app_sec:t2i_samples}, we present the image samples for text-to-image experiment, see~\Cref{sec:experimental_results}. Then, in~\Cref{app_sec:ddim}, we briefly cover the DDIM framework. Following that, in~\Cref{app_sec:other_approaches}, we discuss alternative approaches for learning guidance weights. In~\Cref{app_sec:l2_approach}, we discuss the simple approach based on $\ell_2$ objective~(\eqref{eq:l2_objective}) and in~\Cref{sec_app:learning_to_guide_with_rewards}, we discuss the extension of our method to reward-guided setting. Finally, in~\Cref{app_sec:experimental_details}, we cover all the experimental details and in~\Cref{app_sec:additional_experiments}, we present additional experimental results.

\section{Text to image samples}
\label{app_sec:t2i_samples}

\begin{figure}[h]
\centering
\hfill
\begin{subfigure}{\textwidth}
    \includegraphics[width=0.23\textwidth]{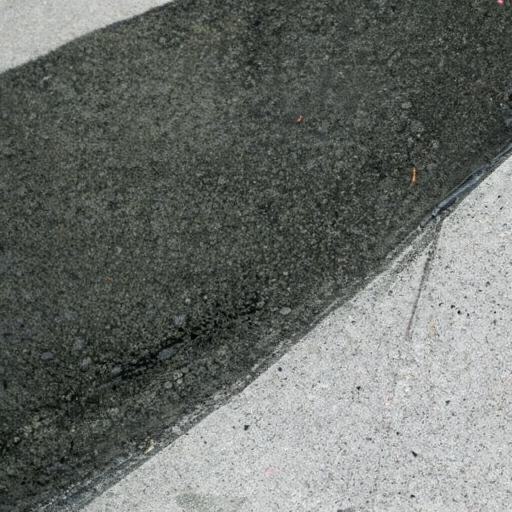}
    \includegraphics[width=0.23\textwidth]{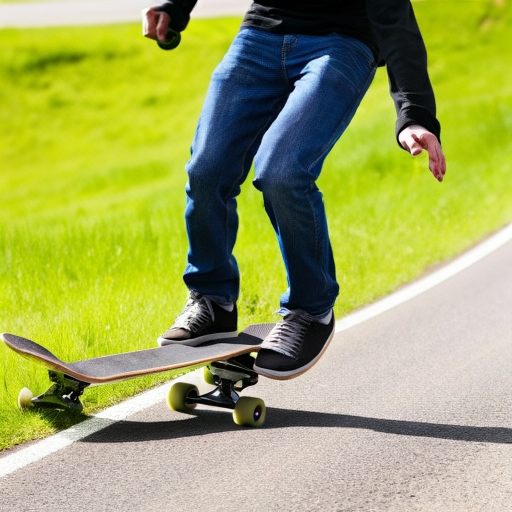}
    \includegraphics[width=0.23\textwidth]{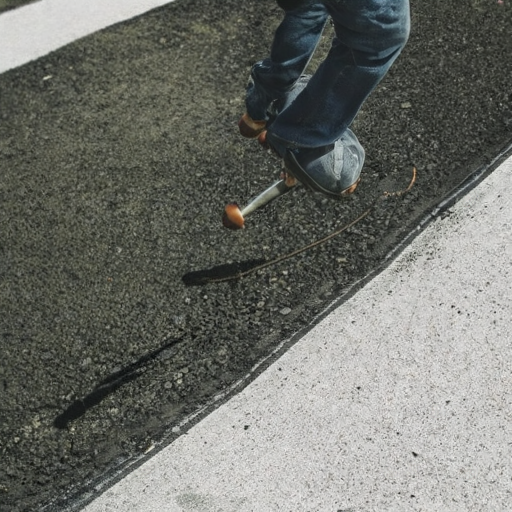}
    \includegraphics[width=0.23\textwidth]{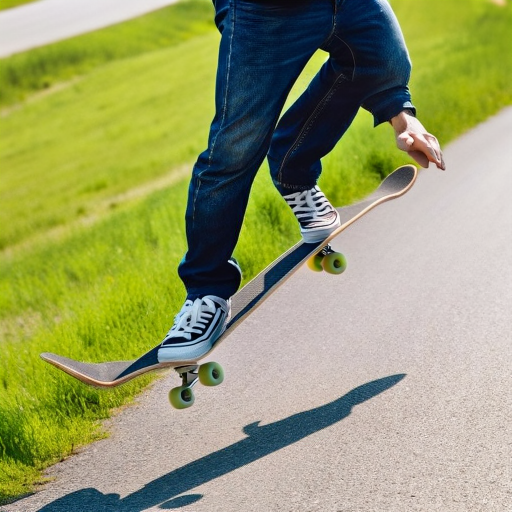}
    \caption{Man performing stunt on a skateboard on a road.}
\end{subfigure}
\hfill
\begin{subfigure}{\textwidth}
    \includegraphics[width=0.23\textwidth]{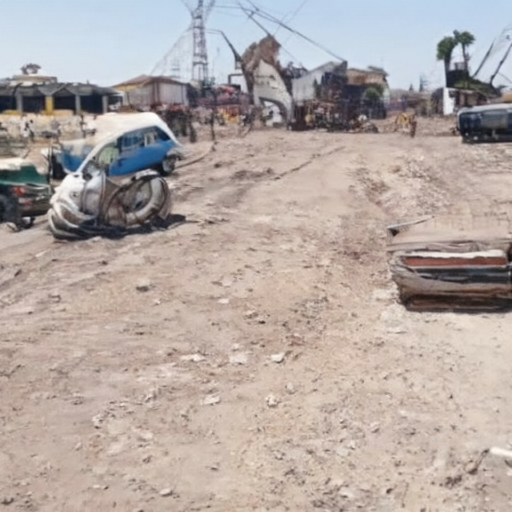}
    \includegraphics[width=0.23\textwidth]{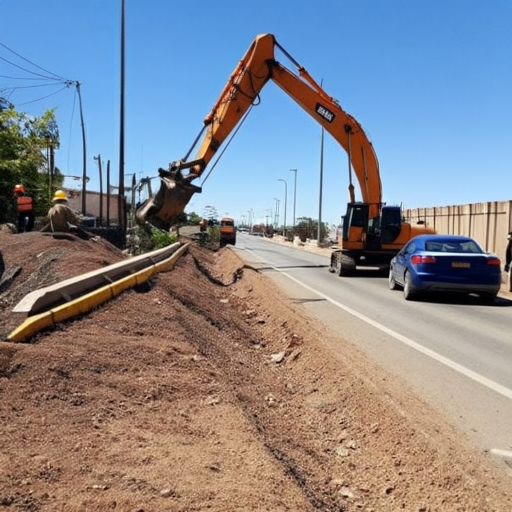}
    \includegraphics[width=0.23\textwidth]{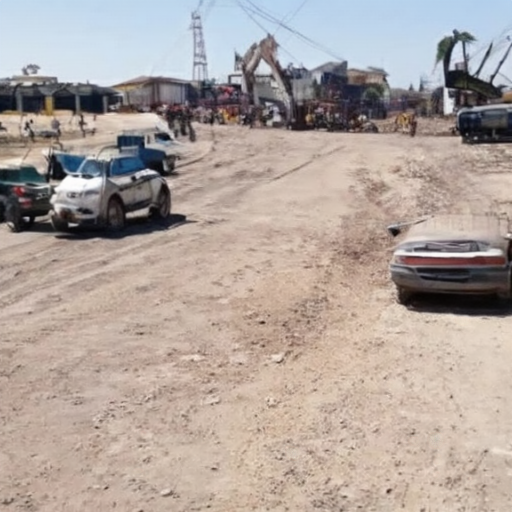}
    \includegraphics[width=0.23\textwidth]{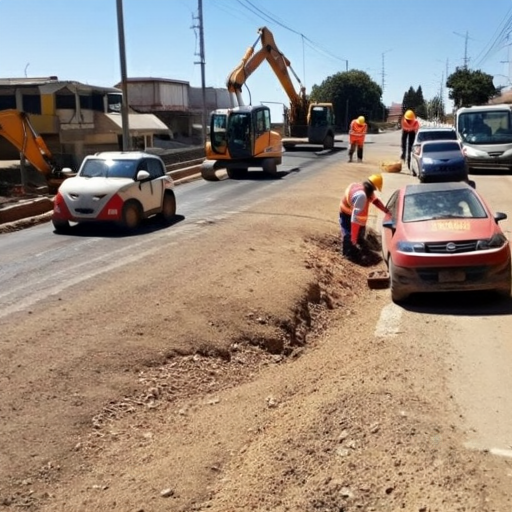}
    \caption{A street with cars and construction workers working.}
\end{subfigure}
\hfill
\begin{subfigure}{\textwidth}
    \includegraphics[width=0.23\textwidth]{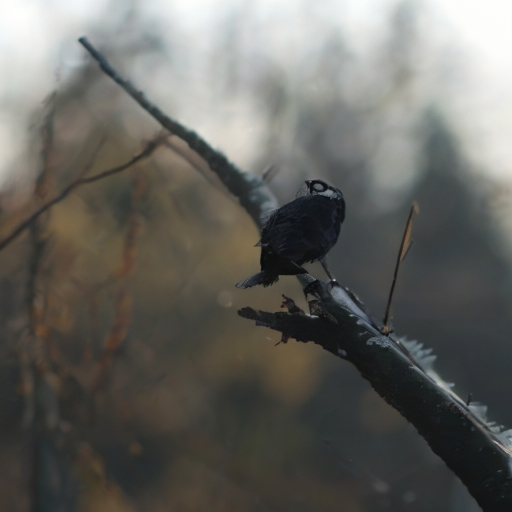}
    \includegraphics[width=0.23\textwidth]{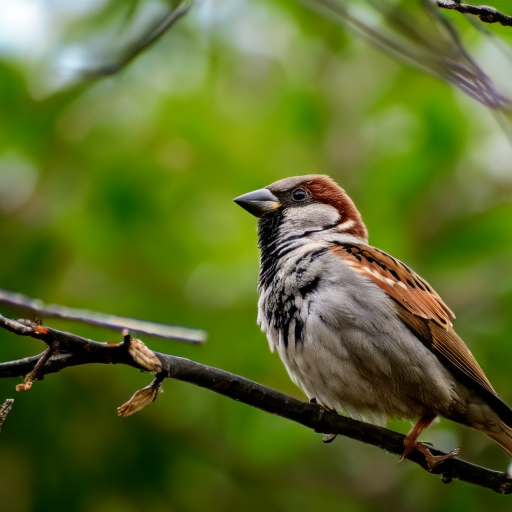}
    \includegraphics[width=0.23\textwidth]{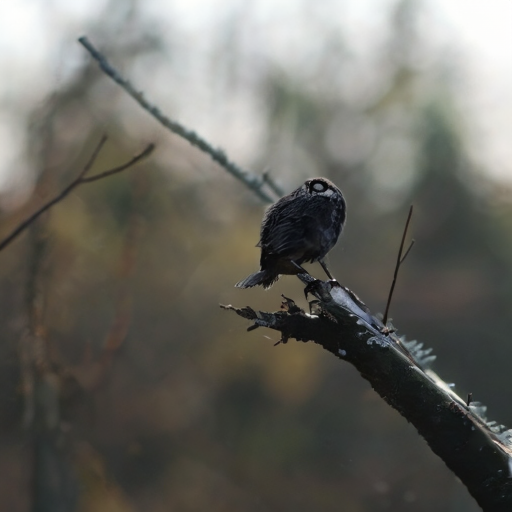}
    \includegraphics[width=0.23\textwidth]{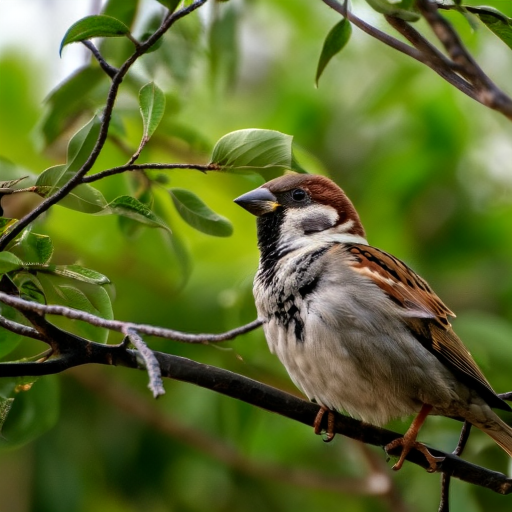}
    \caption{A bird is sitting on a branch among unfocused trees.}
\end{subfigure}
\hfill
\begin{subfigure}{\textwidth}
    \includegraphics[width=0.23\textwidth]{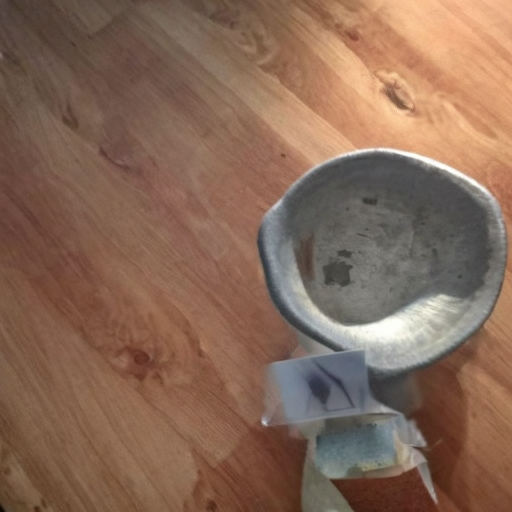}
    \includegraphics[width=0.23\textwidth]{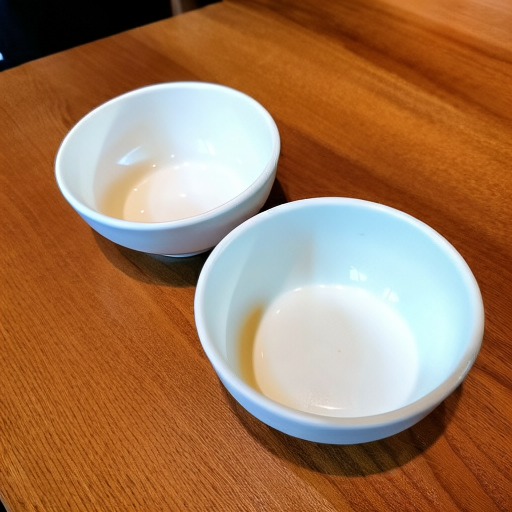}
    \includegraphics[width=0.23\textwidth]{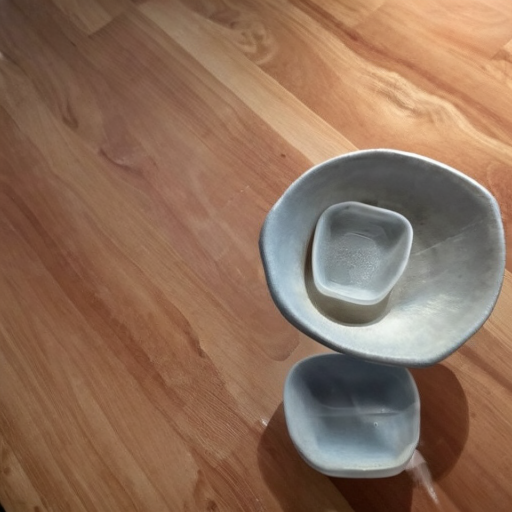}
    \includegraphics[width=0.23\textwidth]{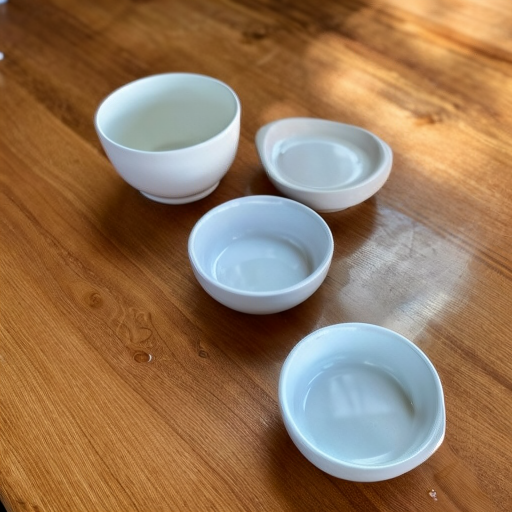}
    \caption{A wooden table topped with four white bowls.}
\end{subfigure}
\hfill
\
\caption{\textbf{T2I Results on MS-COCO. Part 1}. (left-to-right) We provide results of images generated from the given text prompt without CFG, with CFG $\omega = 7.5$, our method with self-consistency loss and our method  with self-consistency loss and CLIP score reward.}
\label{fig:ms-coco-visualization}
\end{figure}

\begin{figure}[h]
\centering
\begin{subfigure}{\textwidth}
    \includegraphics[width=0.23\textwidth]{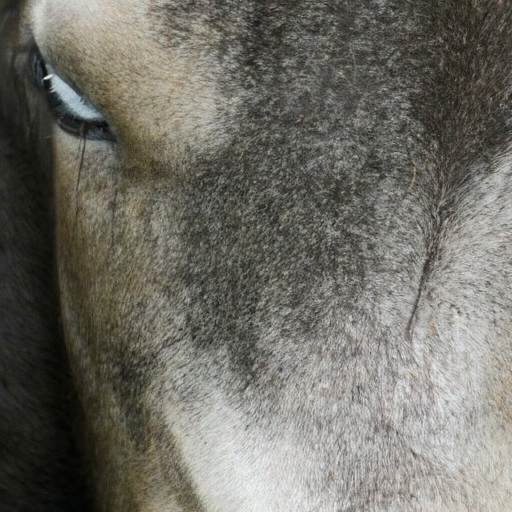}
    \includegraphics[width=0.23\textwidth]{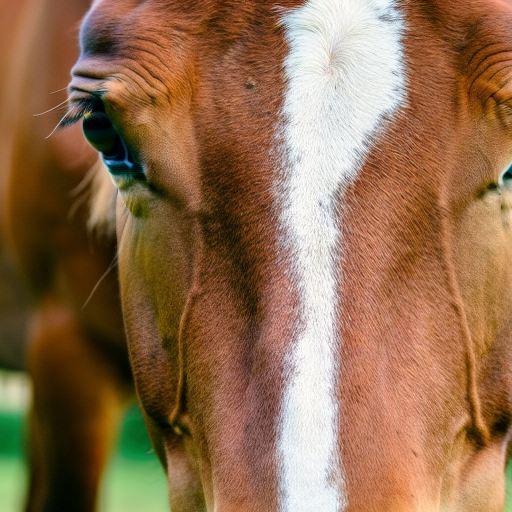}
    \includegraphics[width=0.23\textwidth]{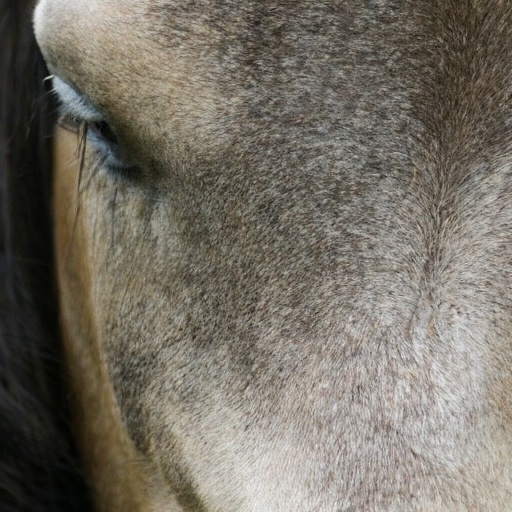}
    \includegraphics[width=0.23\textwidth]{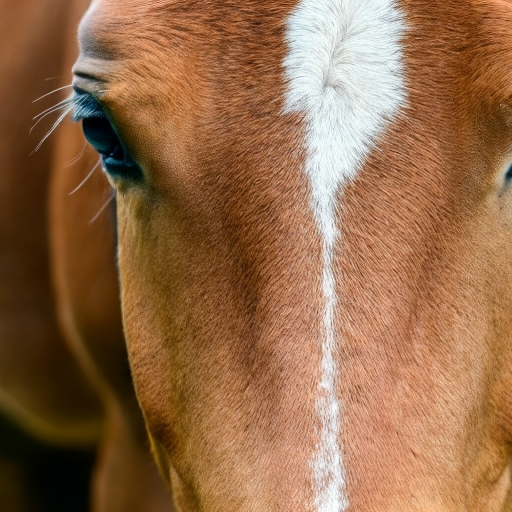}
    \caption{A light brown horse's face is shown at close range.}
\end{subfigure}
\hfill
\begin{subfigure}{\textwidth}
    \includegraphics[width=0.23\textwidth]{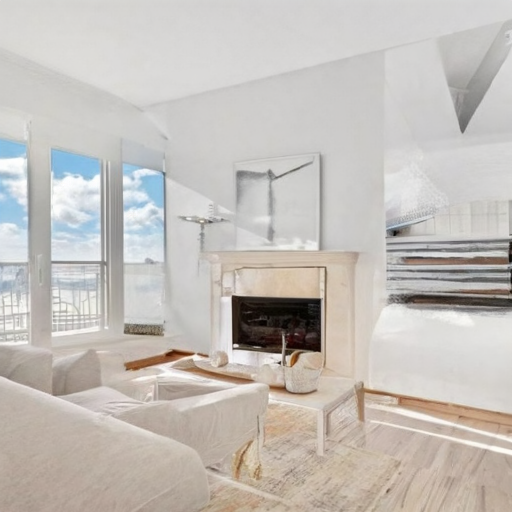}
    \includegraphics[width=0.23\textwidth]{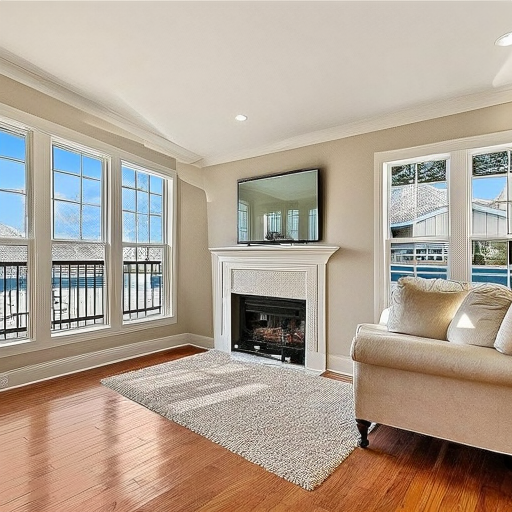}
    \includegraphics[width=0.23\textwidth]{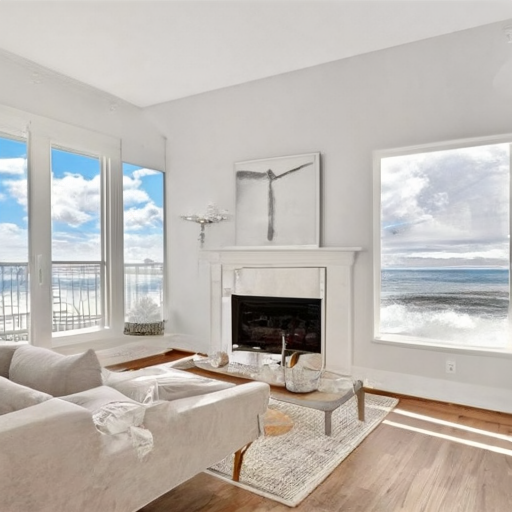}
    \includegraphics[width=0.23\textwidth]{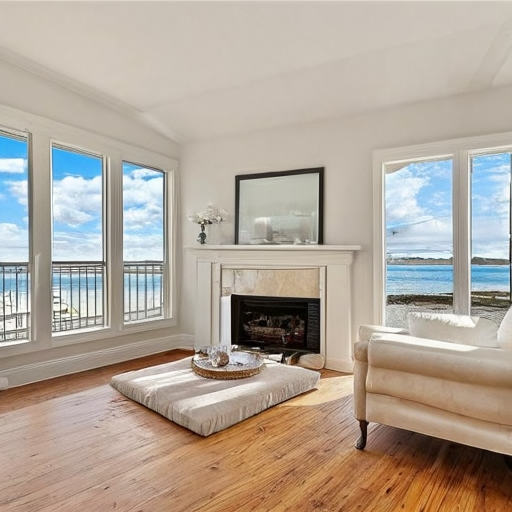}
    \caption{A living room with furniture, a fireplace, and a large scenic window.}
\end{subfigure}
\hfill
\begin{subfigure}{\textwidth}
    \includegraphics[width=0.23\textwidth]{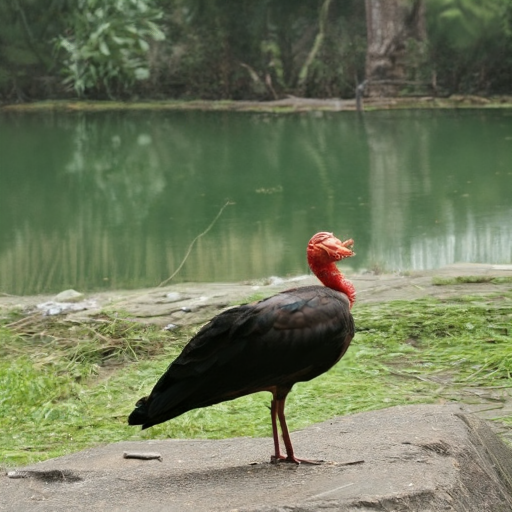}
    \includegraphics[width=0.23\textwidth]{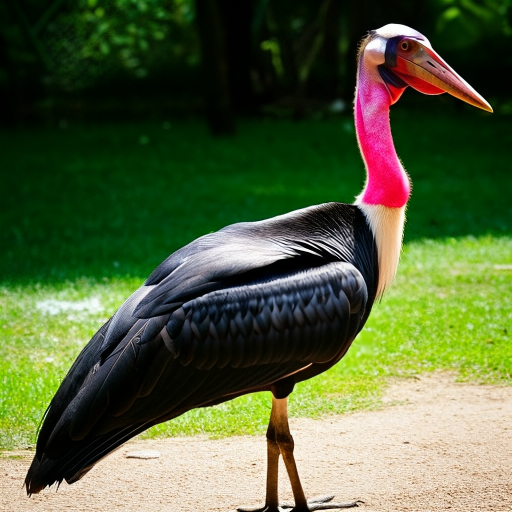}
    \includegraphics[width=0.23\textwidth]{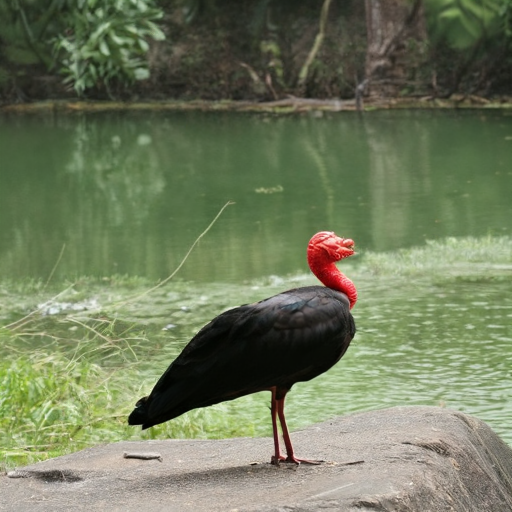}
    \includegraphics[width=0.23\textwidth]{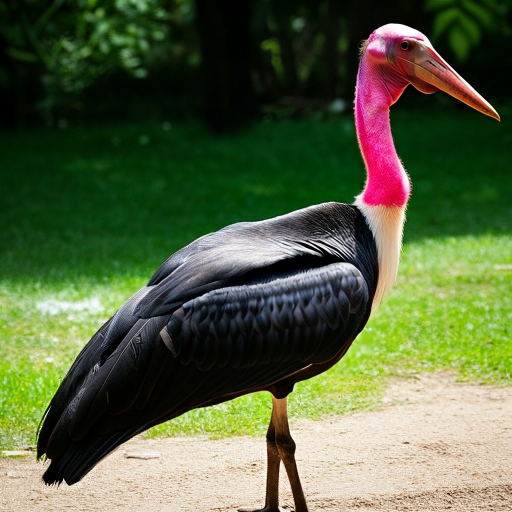}
    \caption{The large bird has a red face and black feathers.}
\end{subfigure}
\hfill
\begin{subfigure}{\textwidth}
    \includegraphics[width=0.23\textwidth]{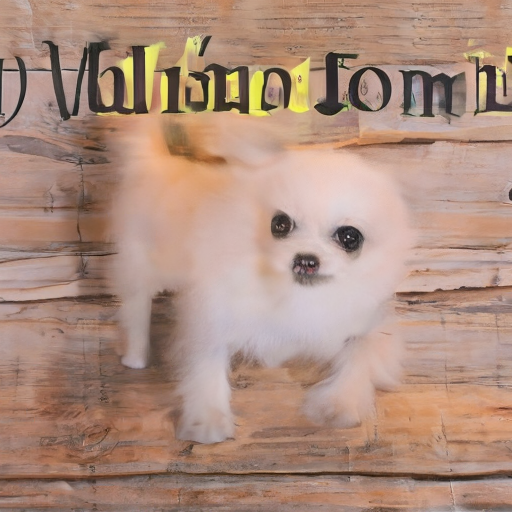}
    \includegraphics[width=0.23\textwidth]{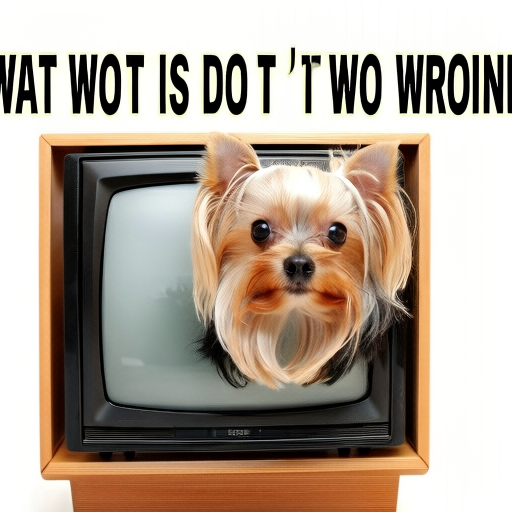}
    \includegraphics[width=0.23\textwidth]{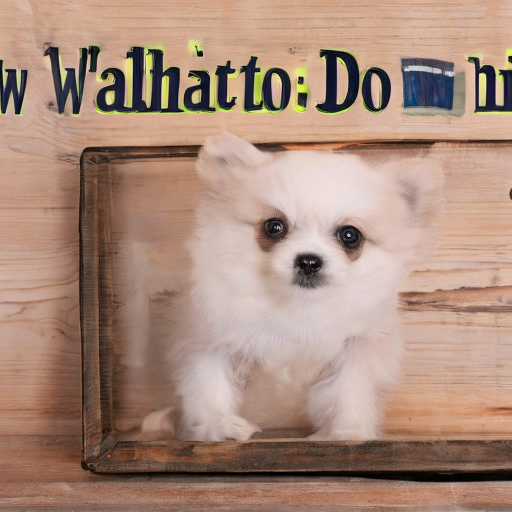}
    \includegraphics[width=0.23\textwidth]{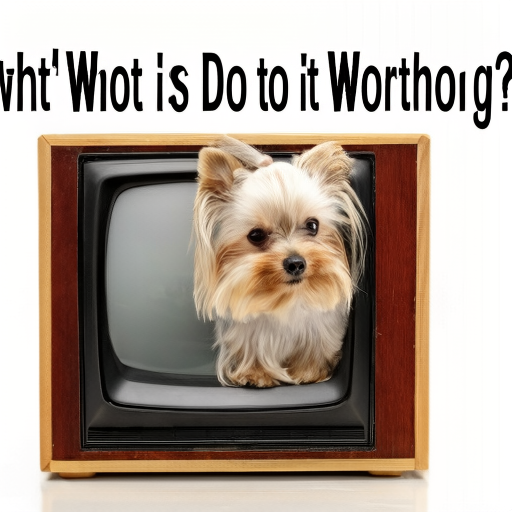}
    \caption{A small dog on TV behind the words " What Did I Do Wrong?}
\end{subfigure}
\hfill
\begin{subfigure}{\textwidth}
    \includegraphics[width=0.23\textwidth]{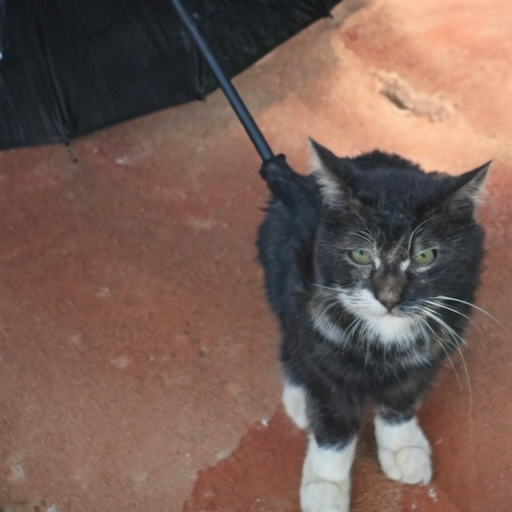}
    \includegraphics[width=0.23\textwidth]{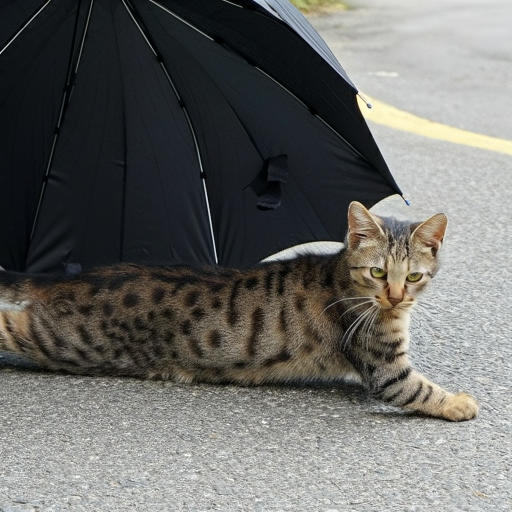}
    \includegraphics[width=0.23\textwidth]{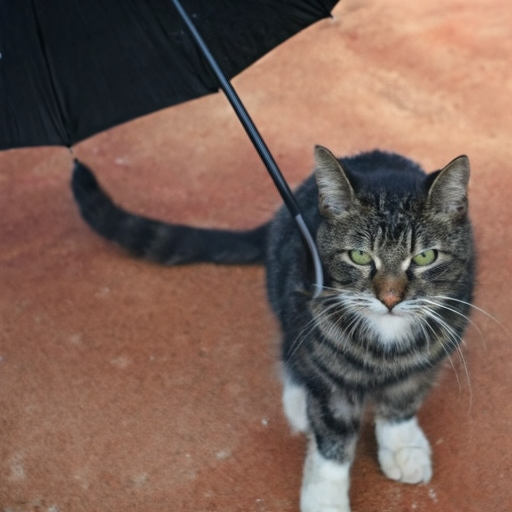}
    \includegraphics[width=0.23\textwidth]{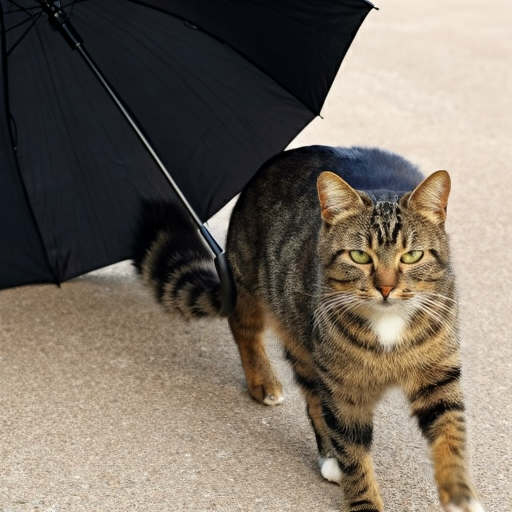}
    \caption{A cat walking under a black open umbrella.}
\end{subfigure}
\caption{\textbf{T2I Results on MS-COCO. Part 2 (Randomly selected)}. (left-to-right) We provide results of images generated from the given text prompt without CFG, with CFG $\omega = 7.5$, our method with self-consistency loss and our method  with self-consistency loss and CLIP score reward.}
\label{fig:ms-coco-visualization-additional}
\end{figure}

\begin{figure}
\centering
\begin{subfigure}{\textwidth}
    \includegraphics[width=0.23\textwidth]{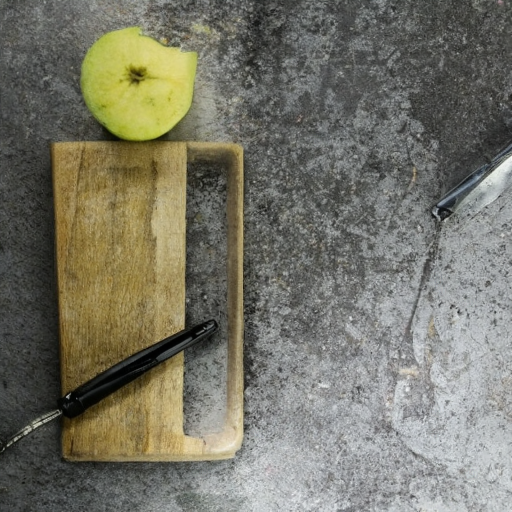}
    \includegraphics[width=0.23\textwidth]{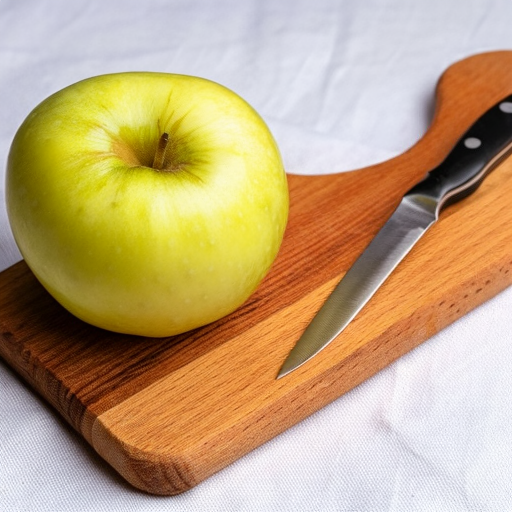}
    \includegraphics[width=0.23\textwidth]{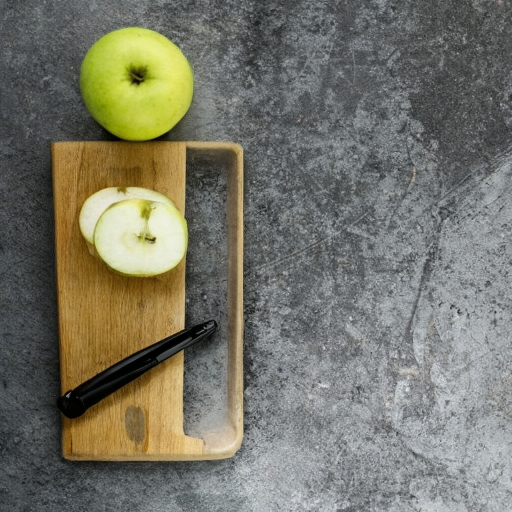}
    \includegraphics[width=0.23\textwidth]{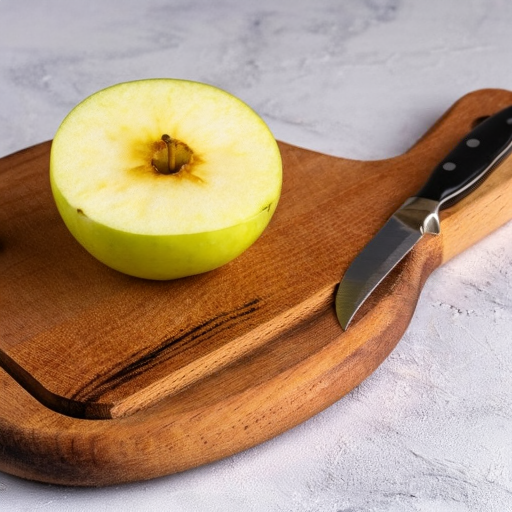}
    \caption{A small cutting board and knife with a cut apple.}
\end{subfigure}
\hfill
\begin{subfigure}{\textwidth}
    \includegraphics[width=0.23\textwidth]{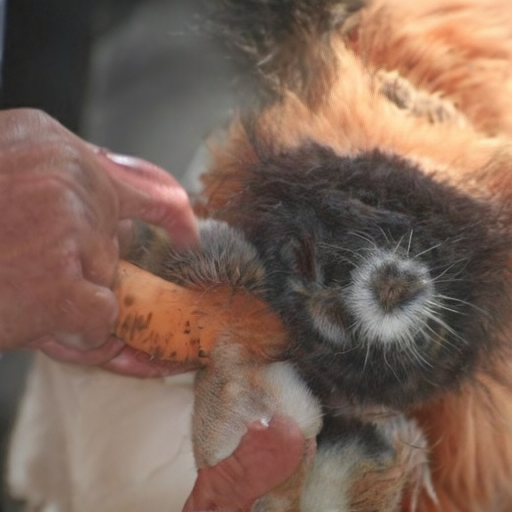}
    \includegraphics[width=0.23\textwidth]{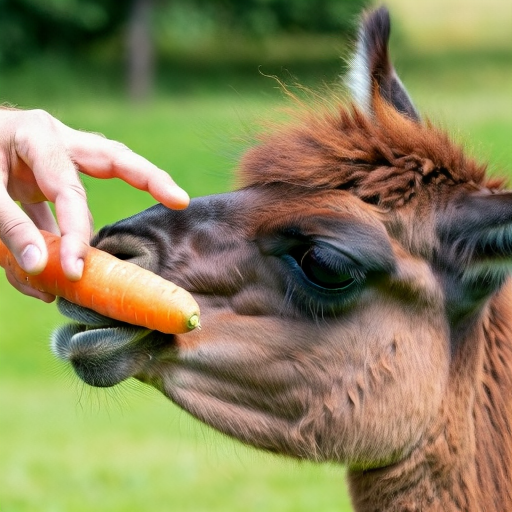}
    \includegraphics[width=0.23\textwidth]{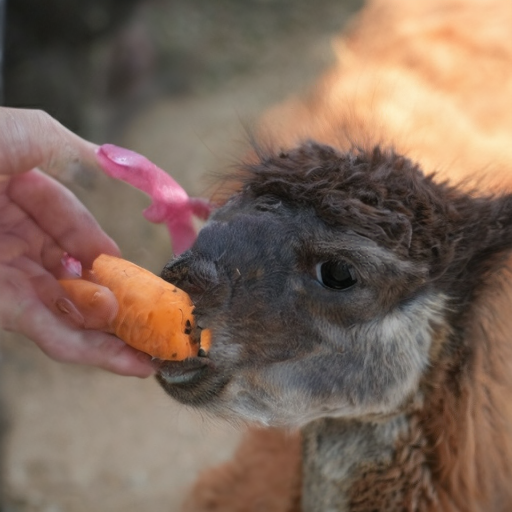}
    \includegraphics[width=0.23\textwidth]{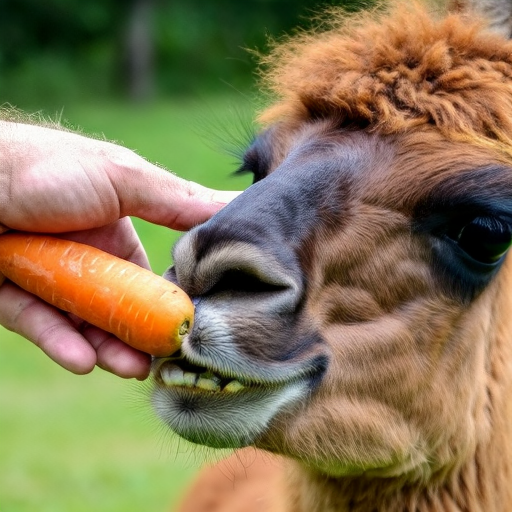}
    \caption{A hand is holding a carrot for a llama to chew.}
\end{subfigure}
\hfill
\begin{subfigure}{\textwidth}
    \includegraphics[width=0.23\textwidth]{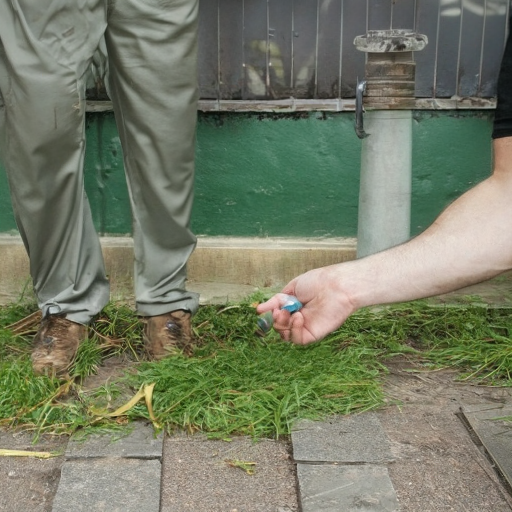}
    \includegraphics[width=0.23\textwidth]{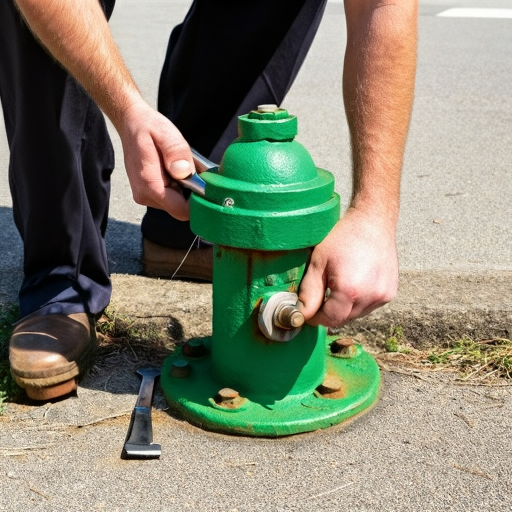}
    \includegraphics[width=0.23\textwidth]{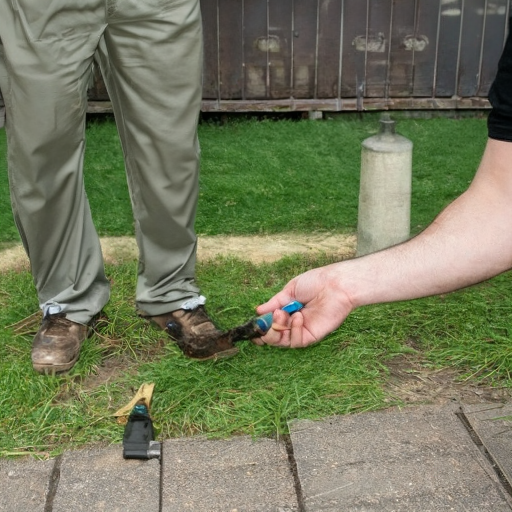}
    \includegraphics[width=0.23\textwidth]{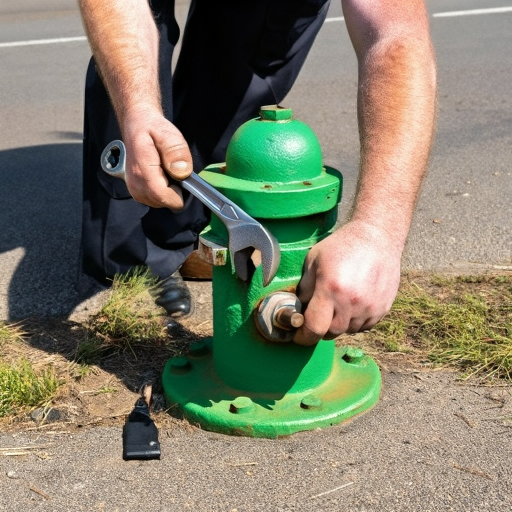}
    \caption{A man with a wrench turning off fire hydrant.}
\end{subfigure}
\hfill
\begin{subfigure}{\textwidth}
    \includegraphics[width=0.23\textwidth]{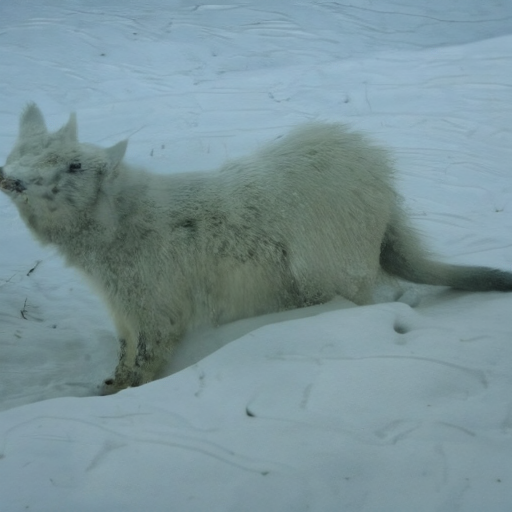}
    \includegraphics[width=0.23\textwidth]{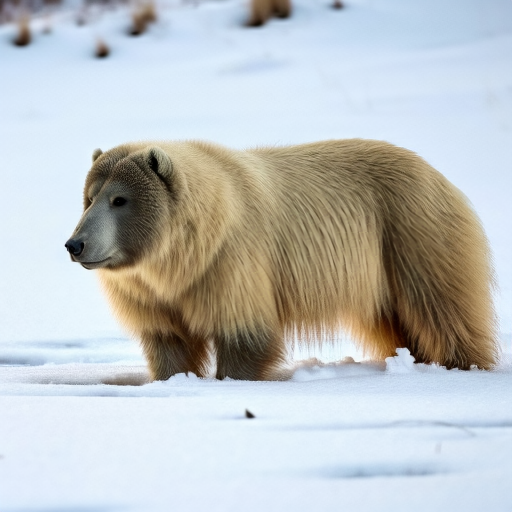}
    \includegraphics[width=0.23\textwidth]{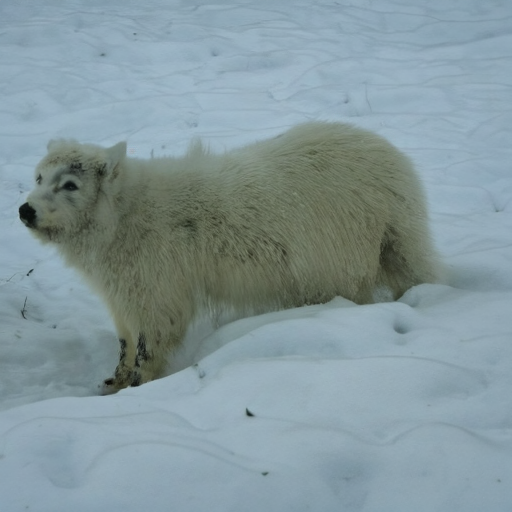}
    \includegraphics[width=0.23\textwidth]{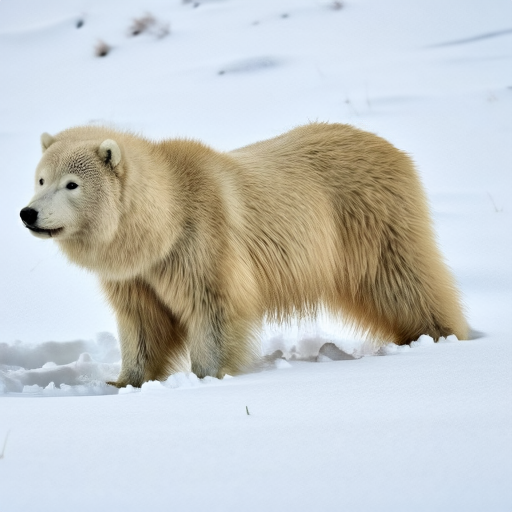}
    \caption{An animal that is in the snow by themselves.}
\end{subfigure}
\hfill
\begin{subfigure}{\textwidth}
    \includegraphics[width=0.23\textwidth]{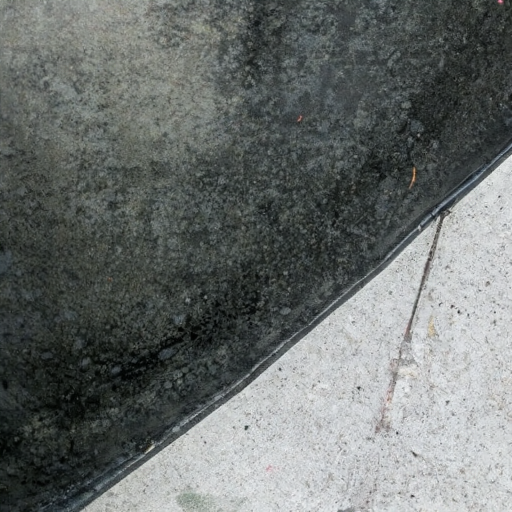}
    \includegraphics[width=0.23\textwidth]{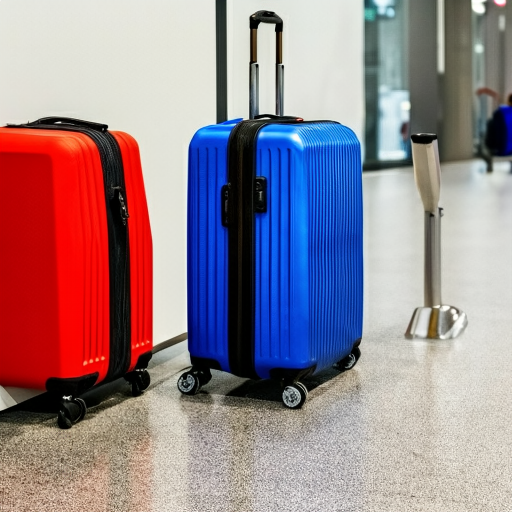}
    \includegraphics[width=0.23\textwidth]{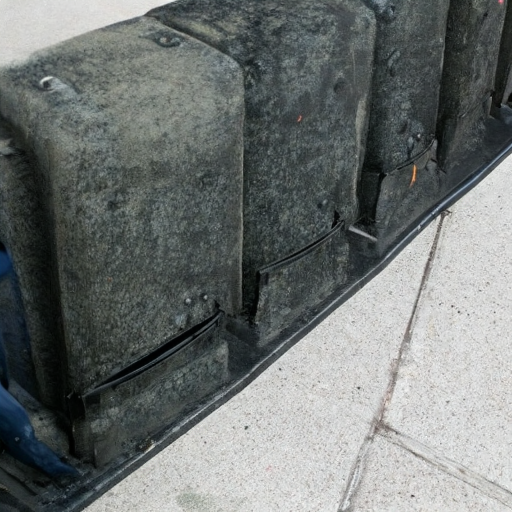}
    \includegraphics[width=0.23\textwidth]{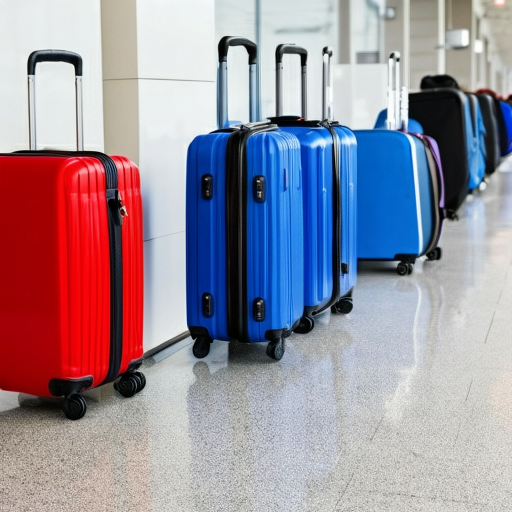}
    \caption{A row of red and blue luggage sitting at an airport.}
\end{subfigure}
\hfill
\caption{\textbf{Additional T2I Results on MS-COCO. Part 3}. (left-to-right) We provide results of images generated from the given text prompt without CFG, with CFG $\omega = 7.5$, our method with self-consistency loss and our method  with self-consistency loss and CLIP score reward.}
\label{fig:ms-coco-visualization-additional_2}
\end{figure}

We provide qualitative results for the text-to-image experiments in Figure~\ref{fig:ms-coco-visualization}, Figure~\ref{fig:ms-coco-visualization-additional} and in Figure~\ref{fig:ms-coco-visualization-additional_2}.

\begin{figure}
\centering
\begin{subfigure}{\textwidth}
    \includegraphics[width=0.23\textwidth]{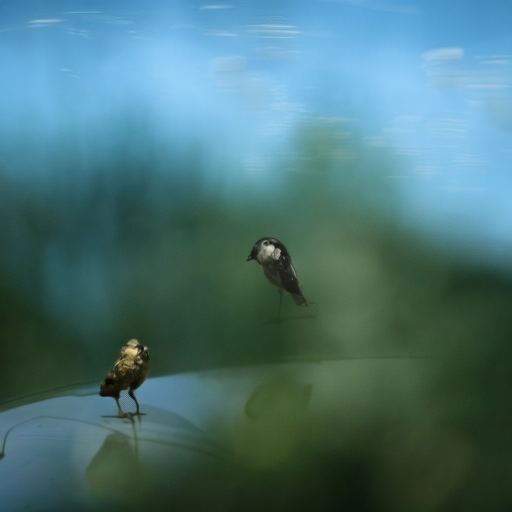}
    \includegraphics[width=0.23\textwidth]{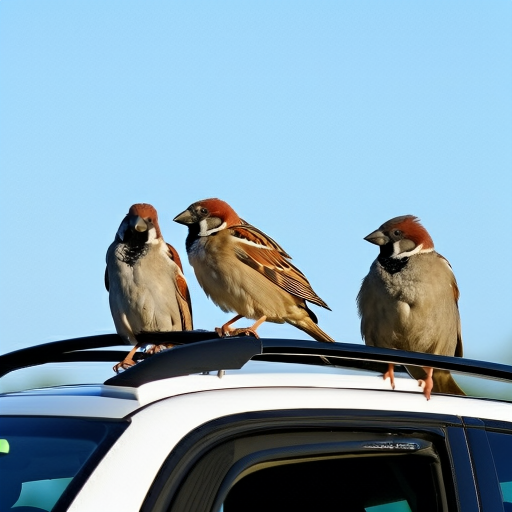}
    \includegraphics[width=0.23\textwidth]{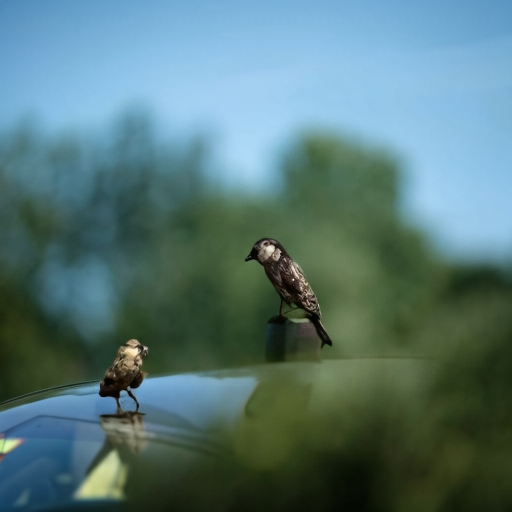}
    \includegraphics[width=0.23\textwidth]{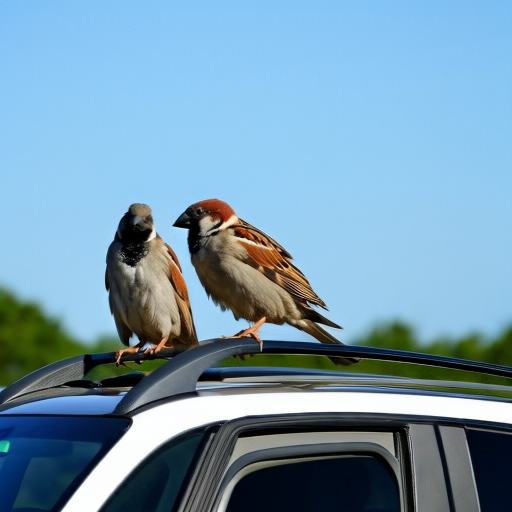}
    \caption{Two birds sit on top of a parked car.}
\end{subfigure}
\hfill
\begin{subfigure}{\textwidth}
    \includegraphics[width=0.23\textwidth]{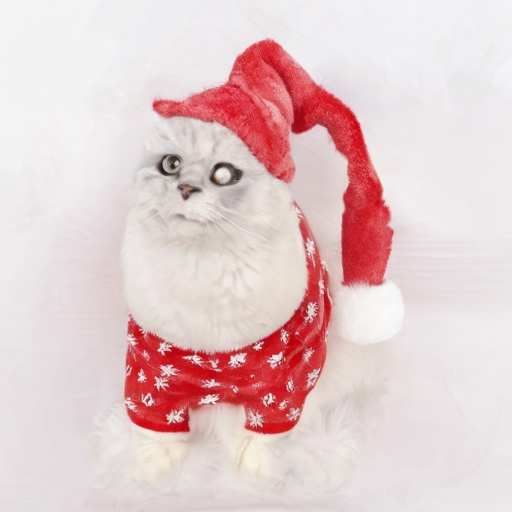}
    \includegraphics[width=0.23\textwidth]{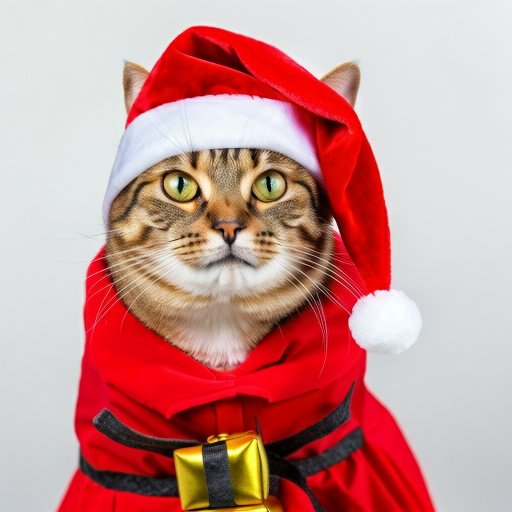}
    \includegraphics[width=0.23\textwidth]{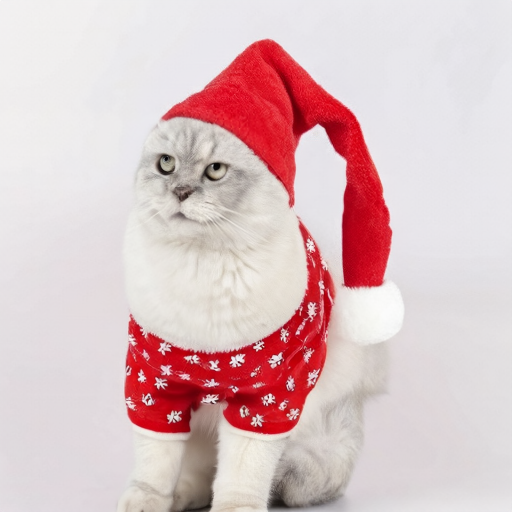}
    \includegraphics[width=0.23\textwidth]{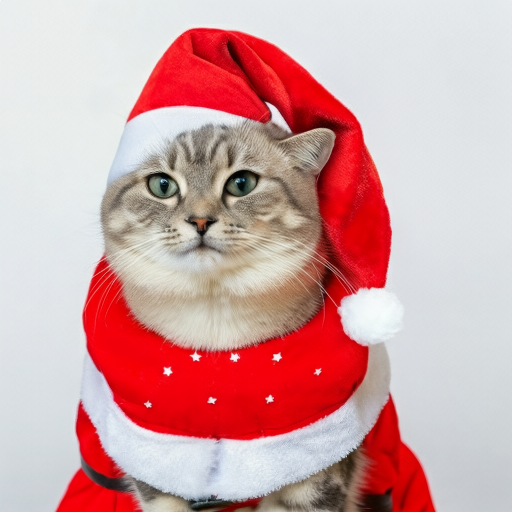}
    \caption{A cat wearing a Santa suit including a Santa hat.}
\end{subfigure}
\hfill
\begin{subfigure}{\textwidth}
    \includegraphics[width=0.23\textwidth]{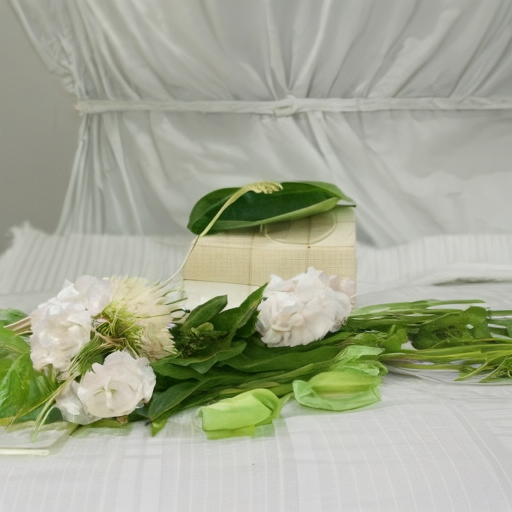}
    \includegraphics[width=0.23\textwidth]{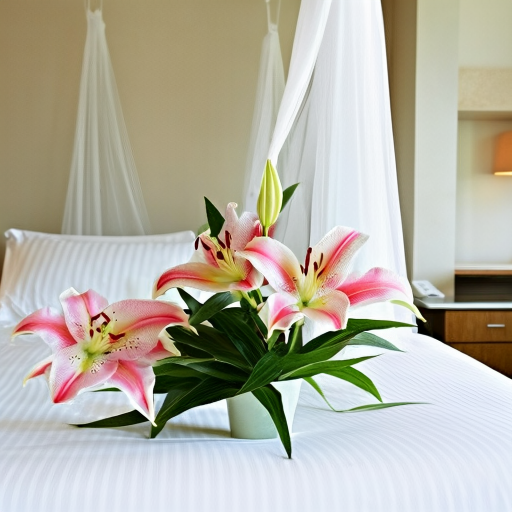}
    \includegraphics[width=0.23\textwidth]{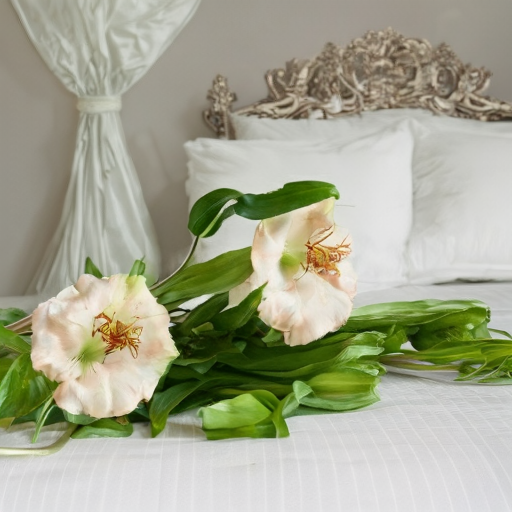}
    \includegraphics[width=0.23\textwidth]{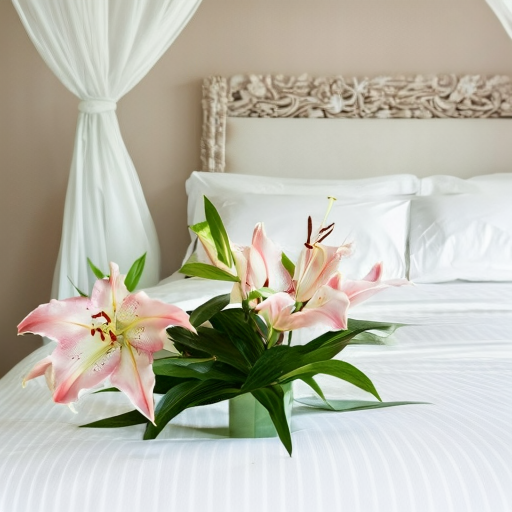}
    \caption{Lilies in a vase are set at the foot of a white canopy bed.}
\end{subfigure}
\hfill
\begin{subfigure}{\textwidth}
    \includegraphics[width=0.23\textwidth]{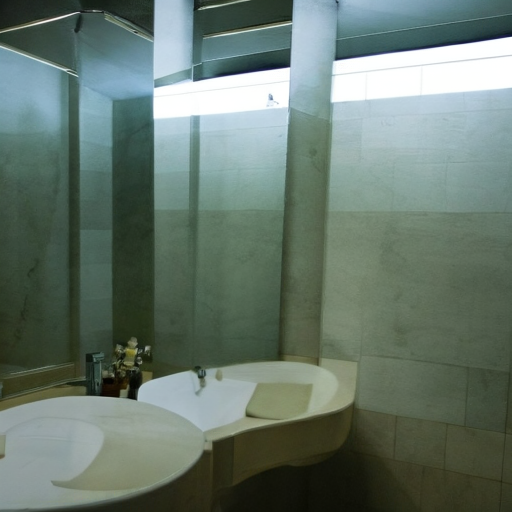}
    \includegraphics[width=0.23\textwidth]{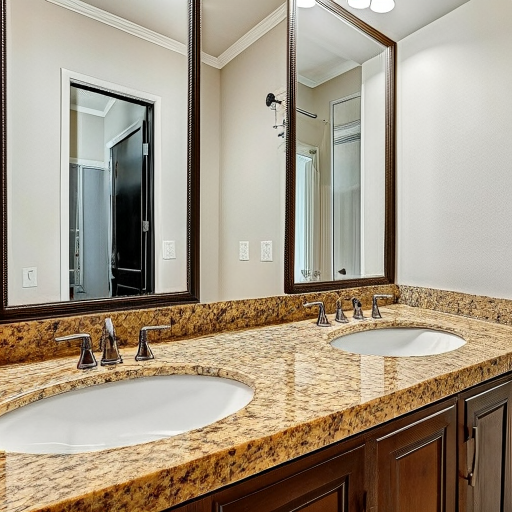}
    \includegraphics[width=0.23\textwidth]{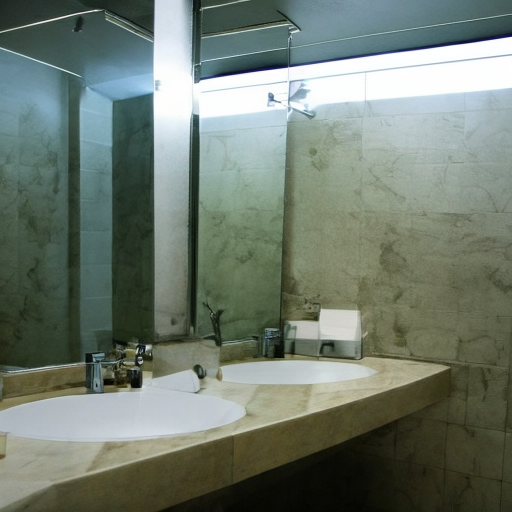}
    \includegraphics[width=0.23\textwidth]{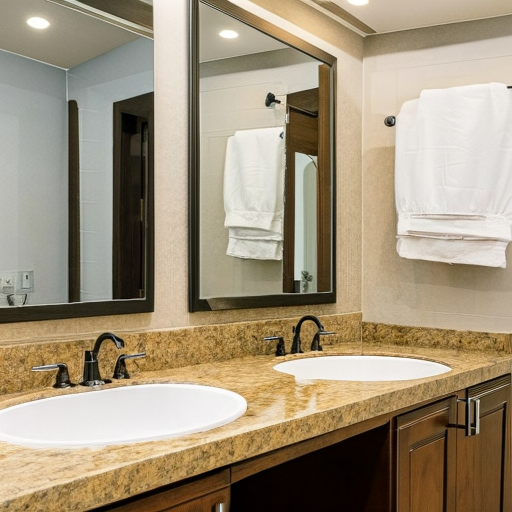}
    \caption{A bathroom with a marble counter under two giant mirrors.}
\end{subfigure}
\hfill
\begin{subfigure}{\textwidth}
    \includegraphics[width=0.23\textwidth]{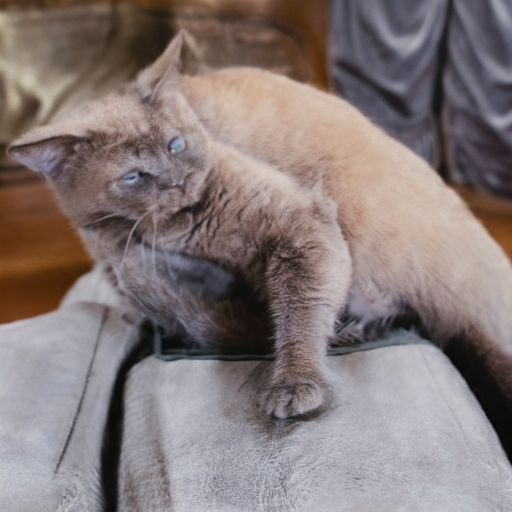}
    \includegraphics[width=0.23\textwidth]{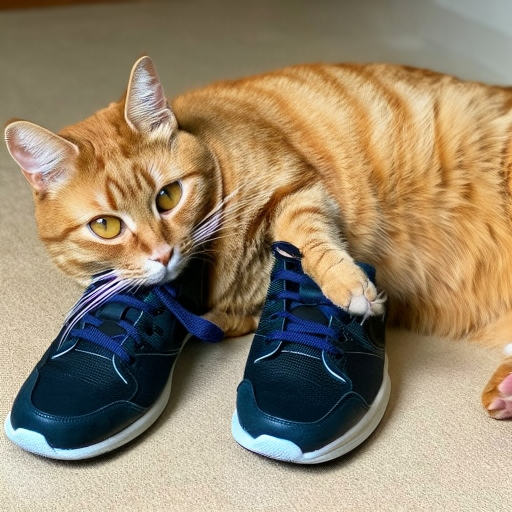}
    \includegraphics[width=0.23\textwidth]{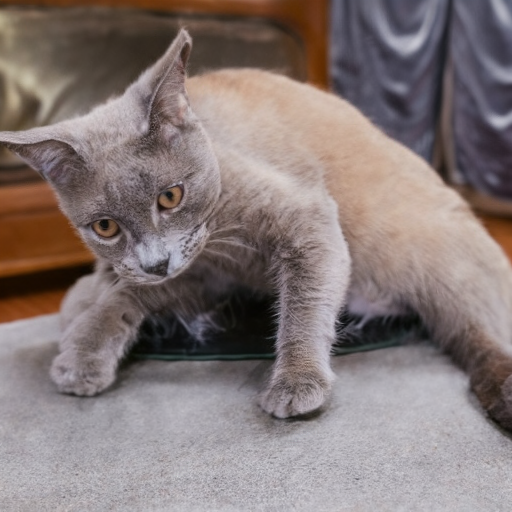}
    \includegraphics[width=0.23\textwidth]{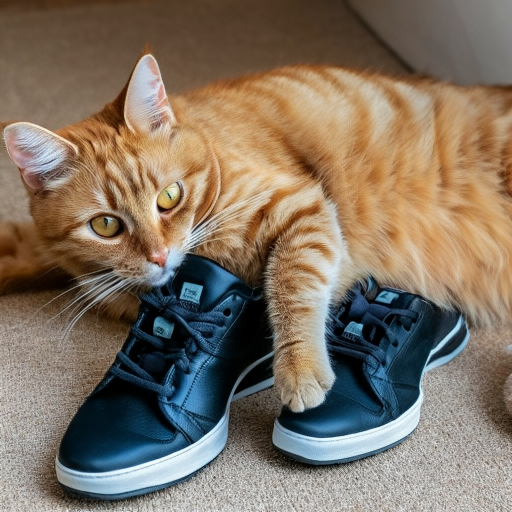}
    \caption{A cat laying on top of shoes on floor.}
\end{subfigure}
\hfill
\caption{\textbf{Additional T2I Results on MS-COCO. Part 4}. (left-to-right) We provide results of images generated from the given text prompt without CFG, with CFG $\omega = 7.5$, our method with self-consistency loss and our method  with self-consistency loss and CLIP score reward.}
\label{fig:ms-coco-visualization-additional_3}
\end{figure}

\section{DDIM framework}
\label{app_sec:ddim}

\paragraph{Diffusion models.}
The goal of conditional diffusion models is to sample from a target conditional distribution $p_0(x|c)$ on $\rset^d$, which we denote as $p_{0|c}$. Here $c$ is some user-specified conditioning signal (i.e., a class label or a text prompt) and $p(c)$ is a conditioning distribution. 

We adopt here the Denoising Diffusion Implicit Models (DDIM) framework of \citet{song2020denoising}. Let $x_{t_0}\sim p_{0|c}$  and define the process $x_{t_1:t_N}:=(x_{t_1},...,x_{t_N})$ by
\begin{equation}\label{eq_app:forward}
   p(x_{t_1:t_N}|x_{t_0})= p(x_{t_N}|x_{t_0}) \prod_{k=1}^{N-1} p(x_{t_{k}} | x_{t_0}, x_{t_{k+1}}) ,
\end{equation}
with $0 = t_0 < \dots < t_N = 1$. For $0\leq s<t\leq 1$, we let
\begin{equation}
    \label{eq_app:transition_density}
    p(x_{s} |x_0, x_{t})=\mathcal{N}(x_{s} ; \mu_{s,t}(x_0, x_{t}), \Sigma_{s,t}),
\end{equation}
The mean and covariance of $p(x_{s} |x_0, x_{t})$ are given by
\begin{align}
    &\mu_{s,t}(x_0, x_t) = (\vareps^2 r_{1,2}(s,t) + (1 - \vareps^2) r_{0,1}) x_{t} + \alpha_{s}(1 - \vareps^2 r_{2,2}(s,t) - (1 - \vareps^2) r_{1,1}(s,t)) x_0 ,
\\
& \Sigma_{s,t} =  \sigma_{s}^2 (1 - (\vareps^2 r_{1,1}(s,t) + (1-\vareps^2))^2) \Id, \label{eq_app:mean_sigma}
\end{align}
with $r_{i,j}(s,t) = \left(\alpha_t/\alpha_s\right)^i \left(\sigma_s^2/\sigma_t^2\right)^j $; see \citep{song2020denoising} and Appendix F in \citep{debortoli2025distributional}. The parameter $\varepsilon \in [0,1]$ in (\eqref{eq_app:mean_sigma}) is a \emph{churn} parameter which interpolates between a \emph{deterministic} process ($\varepsilon=0$) and a \emph{stochastic} one ($\varepsilon=1$). This ensures that for any $t \in [0,1]$, 
\begin{equation}
\label{eq_app:interpolation}
p(x_t | x_0) = \mathcal{N}(x_t ; \alpha_t x_0, \sigma_t^2 \Id) ,
\end{equation}
for $\alpha_t,\sigma_t$ such that $\alpha_0 =\sigma_1 =1$ and $\alpha_1 = \sigma_0 = 0$. In particular, this guarantees that $p(x_1|x_0) = \mathcal{N}(x_1 ; 0,\Id)$.

The data $x_{t_0}\sim p_{0|c}$ is generated by sampling $x_{t_N}\sim \mathcal{N}(0,\Id)$ and 
$x_{t_k}\sim p(\cdot|x_{t_{k+1}})$ for $k=N-1,...,0$, where for $0\leq s<t\leq 1$
\begin{equation}
\label{eq_app:backward_kernel}
    p(x_s|x_t,c) = \int_{\rset^d} p(x_s|x_0, x_t) p(x_0 |c,x_t) \rmd x_0 .
\end{equation}

\begin{algorithm}
\caption{CFG version of DDIM (sampling)}\label{alg:sampling_diffusion_ddim_guided}
\begin{algorithmic}
\Require $\{t_k\}_{k=0}^N$ with $t_0 = 0 < \dots < t_N = 1$, churn parameter $\vareps$, conditioning $c$, guiding weights $\boldsymbol{\omega}=(\omega_{c,(s,t)})$, unconditional and conditional pretrained denoisers $\hat{x}_{\theta}(x_t, \varnothing)$, $\hat{x}_{\theta}(x_t,c)$
\State Sample $x_{t_N} \sim \mathcal{N}(0, \Id)$
\For{$k \in \{N-1, \dots, 0\}$}
\State Sample $z \sim \mathcal{N}(0, \Id)$
\State Set current guidance weight $\omega=\omega_{c,(t_{k},t_{k+1})}$
\State Compute $\hat{x}_0= \hat{x}_{\theta}(t_{k+1}, x_{t_{k+1}},c; \omega)$ using~(\eqref{eq:cfg_delta_term})
\State Compute $\mu_{t_k, t_{k+1}}(\hat{x}_0, x_{t_{k+1}})$,  $\Sigma_{t_k, t_{k+1}}$ using~(\eqref{eq_app:mean_sigma})
\State Sample $x_{t_k} \sim \mathcal{N}(\mu_{t_k, t_{k+1}}(\hat{x}_0, x_{t_{k+1}}), \Sigma_{t_k, t_{k+1}})$
\EndFor
\State \textbf{Return:} $x_0$
\end{algorithmic}
\end{algorithm}

\section{Alternative approaches for learning guidance}
\label{app_sec:other_approaches}

We describe the alternative approaches for learning guidance weights and discuss the their potential downfalls.

\subsection{Guided score matching}
\label{app_sec:guided_score_matching}

The most straightforward approach for learning guidance weights is to simply use guided denoiser~(\eqref{eq:cfg_delta_term}) in the score matching. This leads to the following objective
\begin{equation}
    \label{eq:guided_score_matching}
    \mathcal{L}(\phi) = \int_0^1 \lambda(t) \mathbb{E}_{(x_0,c)\sim p(x_0,c), x_t \sim p_{t|0}(\cdot|x_0)}[\| x_0 - \hat{x}_\theta(x_t, c; \omega^\phi_{c,t}) \|^2] \rmd t,
\end{equation}
where we omit the dependence of the guidance weights on $s$, i.e., $\omega^\phi_{c,(s,t)}=\omega^\phi_{c,t}$. However, this approach is doomed to fail, because at convergence, the solution to~(\eqref{eq:guided_score_matching}) satisfies the following property
\begin{equation}
    \label{eq:guided_score_matching_optimality}
    \hat{x}_\theta(x_t, c; \omega^\phi_{c,t}) \approx \mathbb{E}[x_{0}|x_t,c],
\end{equation}
when $x_t \sim p_{t|0}(x_t|x_0)$ is a sample from noising process~(\eqref{eq:interpolation}). However, the condition~(\eqref{eq:guided_score_matching_optimality}) is also satisfied by the unguided score, when $x_t \sim p_{t|0}(x_t|x_0)$, i.e.
\begin{equation}
    \label{eq_app:unguided_score_solution}
    \hat{x}_\theta(x_t, c) \approx \mathbb{E}[x_{0}|x_t,c].
\end{equation}
By equality of rhs of~(\eqref{eq_app:unguided_score_solution}) and of~(\eqref{eq:guided_score_matching_optimality}), we have $\forall x_t \sim p_{t|0}(x_t|x_0), c \sim p(c), x_0 \sim p_{0|c}(\cdot | c), t \in [0,1]$,
\begin{equation}
    \hat{x}_\theta(x_t, c; \omega^\phi_{c,t}) - \hat{x}_\theta(x_t, c) = \omega^\phi_{c,t} \left(\hat{x}_\theta(x_t, c) - \hat{x}_\theta(x_t, \varnothing) \right) \approx 0
\end{equation}
We then assume that $\Delta_{\theta}(x_t,c)=\hat{x}_\theta(x_t, c) - \hat{x}_\theta(x_t, \varnothing) \neq 0$. It is a reasonable assumption since if $\Delta_{\theta}(x_t,c) = 0$, the CFG would not have had any effect. This implies that
\begin{equation}
    \label{eq_app:guided_sm_weights}
    \omega^\phi_{c,t} = 0, \quad \forall c \sim p(c), t \in [0,1]
\end{equation}
The only regime where~(\eqref{eq_app:guided_sm_weights}) will not hold is when
\begin{equation}
    \hat{x}_\theta(x_t, c) \not\approx \mathbb{E}[x_{0}|x_t,c]
\end{equation}
In such a case, one could still find guidance weights for~(\eqref{eq:guided_score_matching_optimality}) to hold. But this is a regime of a very under-trained model. In our image experiments (see~\Cref{sec:experimental_results}), we initially tried using this approach and found that it was leading to zero guidance weights~(\eqref{eq_app:guided_sm_weights}).

\paragraph{Connection to backward $\KL$.} Another explanation for the behavior above is that as we train~(\eqref{eq:guided_score_matching}), we only observe $x_t \sim p(x_t | x_0)$. The form of~(\eqref{eq:guided_score_matching}) comes from optimizing backward $\KL$
\begin{equation}
    \label{eq_app:backwards_kl}
    \KL [p_{[0,1]} || p_{[0,1]}^{\theta,\boldsymbol{\omega}}]  = \int_{0}^{1} \int \int ||x_0 - \hat{x}_{\theta}(x_t,c; \boldsymbol{\omega})||^2 p(x_t | x_0) p(x_0|c) \rmd x_0 \rmd x_t \rmd t,
\end{equation}
where $p_{[0,1]}^{\theta,\boldsymbol{\omega}}$ is the distribution of all the trajectories obtained by running guided backwards diffusion, and $p_{[0,1]}$ is the distribution of all the trajectories from the true diffusion process.

When we use CFG during sampling, we actually may observe noisy samples $\hat{x}_t$ which do not coincide with any $x_{t} \sim p_{t|0}$ form the noising process~(\eqref{eq:interpolation}). This hints that instead of backwards $\KL$~(\eqref{eq_app:backwards_kl}), we should actually aim to optimize the forward $\KL$
\begin{equation}
    \label{eq_app:forward_kl}
    \KL [ p_{[0,1]}^{\theta,\boldsymbol{\omega}} || p_{[0,1]}].
\end{equation}
An approach based on optimizing~(\eqref{eq_app:forward_kl}) was considered in \citet{azangulov2025adaptivediffusionguidancestochastic}. This approach, however, is very computationally expensive since it requires sampling full guided trajectory from the guided backwards process and then compute the gradients on it.

\subsection{Conditional consistency}
\label{sec_app:conditional_consistency}
Here, we describe an approach similar to \emph{marginal-consistency}~(\eqref{eq:marginal_consistency}), but which however could be more effective in practice in principle due to lower variance. We fix conditioning $c$ and we denote by
\begin{equation}
    \label{eq_app:guided_conditional}
    p_{s|0,c}^{t,(\theta,\boldsymbol{\omega})}(x_s | x_0, c) = \int p_{s|t,c}^{(\theta,\boldsymbol{\omega})}(x_s|x_t, c) p_{t|0}(x_t|x_0) \rmd x_t,
\end{equation}
which is the term under integral in~(\eqref{eq:guided_backwards}) depending on $(x_0,c)$. We now consider a \emph{conditional} consistency condition where we fix $c$ and marginalize over $x_0$
\begin{equation}
    \label{eq_app:class_conditional_consistency}
    \int p_{s|0,c}^{t,(\theta,\boldsymbol{\omega})}(x_s | x_0, c) p_{0|c}(x_0 | c) \rmd x_0 = p_{s|c}^{t,(\theta,\boldsymbol{\omega})}(x_s|c) \approx p_{s|c}(x_s|c) = \int p_{s|0}(x_s|x_0) p_{0|c}(x_0 | c) \rmd x_0.
\end{equation}
The main difference with~(\eqref{eq:marginal_consistency}) is that we fixed conditioning $c$. We could use a similar loss
\begin{equation}
    \label{eq_app:conditional_objective}
    \mathcal{L}_c(\boldsymbol{\omega}) = \mathbb{E}_{c \sim p_c, (s, t) \sim p(s,t)} [\MMD_{(\beta,\lambda)}(p_{s|c}^{t,(\theta,\boldsymbol{\omega})}(\cdot|c), p_{s|c}(\cdot|c))],
\end{equation}

which can be optimized by drawing conditional samples $x_0|c$. While this approach is expected to have a lower variance than~(\eqref{eq:marginal_objective}), since we do not need to sample $c$, it requires to be able to produce conditional samples which is not always available in practice during training (i.e., in text-to-image applications). In practice, we also found that it did not lead to good results. We hypothesize that it is due to still quite a high variance of the gradients of $\mathcal{L}_c(\boldsymbol{\omega})$.

\section{L2 approach for learning guidance}
\label{app_sec:l2_approach}

We describe here the algorithm based on the $\ell_2$ objective function~(\eqref{eq:l2_objective})
\begin{align}
    \mathcal{L}_{\ell_2}(\boldsymbol{\omega}) &=\mathbb{E}_{(x_0, c) \sim p_{0,c}, s,t \sim p(s,t)} \left[
    \mathbb{E}[||\tilde{x}_s(\boldsymbol{\omega}) - x_s||^2_2]
    \right]\\
    &=\mathbb{E}_{(x_0, c) \sim p_{0,c}, s,t \sim p(s,t)} \left[\left\|\mathbb{E}[\tilde{x}_s(\boldsymbol{\omega})]-\mathbb{E}[x_s]\right\|^2+\textup{Trace}(\textup{Cov}(\tilde{x}_s(\boldsymbol{\omega})))+\textup{Trace}(\textup{Cov}(x_s))\right]
\end{align}
So, by minimizing this loss function, we attempt to match on averages the means of $\tilde{x}_s(\boldsymbol{\omega})$ and $x_s$ while minimizing the expected sum of variances of the component of $\tilde{x}_s(\boldsymbol{\omega})$.
\paragraph{Empirical objective.} We sample $\{x^i_0,c^i\}_{i=1}^n \overset{\mathrm{i.i.d.}}{\sim} p_{0,c}$ from the training set and we additionally sample noise levels $\{s_i\}_{i=1}^{n} \overset{\mathrm{i.i.d.}}{\sim} \mathcal{U}[S_{\min},1-\zeta- \delta]$, as well as time increments $\{\Delta t_{i}\}_{i=1}^n \overset{\mathrm{i.i.d.}}{\sim} \mathcal{U}[\delta, 1-\zeta -s_i]$ and we let $t_i = s_i + \Delta t_{i}$. For each $(x^i_0, s_i)$, we sample $x_{s_{i}} \sim p_{s_i | 0}(\cdot | x_0^i)$ from the \emph{noising process}~(\eqref{eq:interpolation}), which defines the target samples. We then also produce $\tilde{x}_{s_{i}}(\boldsymbol{\omega}^\phi) \sim p_{s_i|0, c_i}^{t_i, (\theta,\boldsymbol{\omega}^\phi)}(\cdot|x_0^i, c_i)$ by first sampling $\tilde{x}_{t_{i}} \overset{\mathrm{i.i.d.}}{\sim}  p_{t_{i}| 0}(\cdot | x_0^i)$ from the \emph{noising process}~(\eqref{eq:interpolation}) and then denoising with guidance and DDIM using~(\eqref{eq:guided_forward_sample_2}). This defines the proposal samples. We expand loss function~(\eqref{eq:l2_objective}) as a function of guidance network parameters $\phi$ defined on the empirical batches as follows
\begin{equation}
    \label{eq:empirical_l2_loss}
    \hat{\mathcal{L}}_{\ell_2}(\phi) = \frac{1}{n}\sum_{i=1}^{n}\left[ ||\tilde{x}_{s_{i}}(\boldsymbol{\omega}^\phi) - x_{s_{i}}||_{2}^2 \right],
\end{equation}
The full algorithm is given in~\Cref{app_alg:l2_algorithm}.

\begin{algorithm}
\caption{Learning to Guide with $\ell_2$ loss}
\begin{algorithmic}[1] % The [1] enables line numbers
    \State \textbf{Input:} Init. guidance parameters $\phi$; (frozen) denoiser $\hat{x}_\theta$; data distribution $p_0$; learning rate $\eta$; $\zeta > 0$, $S_{\min}>0$, $\delta>0$, b.s. $n$, DDIM churn $\varepsilon \in [0,1]$.
    \Repeat
        \State Sample batch of clean data and their conditionings $\{x^i_0,c^i\}_{i=1}^n \overset{\mathrm{i.i.d.}}{\sim} p_{0,c}$.
        \State Sample $\{s_i\}_{i=1}^{n} \overset{\mathrm{i.i.d.}}{\sim} \mathcal{U}[S_{\min},1-\zeta- \delta]$, $\{\Delta t_{i} \}_{i=1}^n \overset{\mathrm{i.i.d.}}{\sim} \mathcal{U}[\delta, 1-\zeta-s_i]$, let $t_i=s_i + \Delta t_i$
        \State (\textbf{True process}) Sample $x_{s_{i}}  \sim p_{s_i | 0}(\cdot | x_0^i)$ from noising process~(\eqref{eq:interpolation})
        \State (\textbf{Guided process}) Sample  $\tilde{x}_{t_{i}} \sim p_{t_{i}| 0}(\cdot | x_0^i)$ from noising process~(\eqref{eq:interpolation})
        \State Compute guidance weights $\omega_{i} = \omega^\phi_{c_i, (s_i,t_i)}$
        % (s_i, t_{i}, c_i; \phi)$
        and $\hat{x}_{\theta}(\tilde{x}_{t_i}, c_i; \omega_i)$ using~(\eqref{eq:cfg_delta_term})
        \State Sample $\tilde{x}_{s_i}(\boldsymbol{\omega}^\phi)\sim p_{s_i|t_i,0}(\cdot |\tilde{x}_{t_{i}}, \hat{x}_{\theta}(\tilde{x}_{t_i}, c_i; \omega_i))$ from DDIM~(\eqref{eq:transition_density}) with \emph{churn} parameter $\varepsilon$
        \State (\textbf{Loss}) Compute loss
        \State $\hat{\mathcal{L}}_{\ell_2}(\phi) = \frac{1}{n}\sum_{i=1}^{n}\left[ ||\tilde{x}_{s_{i}}(\boldsymbol{\omega}^\phi) - x_{s_{i}}||_{2}^2 \right]$
        \State Update $\phi \leftarrow \phi - \eta \nabla_{\phi} \hat{\mathcal{L}}_{\beta,\lambda}(\phi)$
    \Until{convergence}
    \State \textbf{Output:} Optimized guidance network parameters $\phi$
\end{algorithmic}
\label{app_alg:l2_algorithm}
\end{algorithm}

\section{Learning to guide with rewards}
\label{sec_app:learning_to_guide_with_rewards}

In this section, we describe how our guidance learning approach is extended to a setting, where a reward function $R(x_0,c)$ is available, see~\Cref{alg:learning_to_guide_with_rewards} below.

\begin{algorithm}
\caption{Learning to Guide with reward}
\begin{algorithmic}[1] % The [1] enables line numbers
    \State \textbf{Input:} Init. guidance parameters $\phi$; (frozen) denoiser $\hat{x}_\theta$; data distribution $p_0$; learning rate $\eta$; $\zeta > 0$, $S_{\min}>0$, $\delta>0$, b.s. $n$, n. of particles $m$, $\lambda \in [0,1]$,$\beta \in [0,2]$, DDIM churn $\varepsilon \in [0,1]$. Reward function $R(x_0, c)$, reward loss weight $\gamma_{R} \geq 0$.
    \Repeat
        \State Sample batch of clean data and their conditionings $\{x^i_0,c^i\}_{i=1}^n \overset{\mathrm{i.i.d.}}{\sim} p_{0,c}$.
        \State Sample $\{s_i\}_{i=1}^{n} \overset{\mathrm{i.i.d.}}{\sim} \mathcal{U}[S_{\min},1-\zeta- \delta]$, $\{\Delta t_{i} \}_{i=1}^n \overset{\mathrm{i.i.d.}}{\sim} \mathcal{U}[\delta, 1-\zeta-s_i]$, let $t_i=s_i + \Delta t_i$
        \State (\textbf{True process}) Sample $m$ particles $\{x_{s_{i}}^{j}\}_{j=1}^m  \overset{\mathrm{i.i.d.}}{\sim} p_{s_i | 0}(\cdot | x_0^i)$ from noising process~(\eqref{eq:interpolation})
        \State (\textbf{Guided process}) Sample $m$ particles $\{\tilde{x}^j_{t_{i}}\}_{j=1}^m \sim p_{t_{i}| 0}(\cdot | x_0^i)$ from noising process~(\eqref{eq:interpolation})
        \State Compute guidance weights $\omega_{i} = \omega^\phi_{c_i, (s_i, t_i)}$
        % (s_i, t_{i}, c_i; \phi)$
        and $\hat{x}_{\theta}(\tilde{x}^j_{t_i}, c_i; \omega_i)$ using~(\eqref{eq:cfg_delta_term})
        \State Sample $\tilde{x}_{s_i}^j(\boldsymbol{\omega}^\phi)\sim p_{s_i|t_i,0}(\cdot |\tilde{x}^j_{t_{i}}, \hat{x}_{\theta}(\tilde{x}^j_{t_i}, c_i; \omega_i))$ from DDIM~(\eqref{eq:transition_density}) with \emph{churn} parameter $\varepsilon$
        \State (\textbf{Loss}) Compute loss
        \State $\hat{\mathcal{L}}_{\beta,\lambda}(\phi) = \frac{1}{n}\sum_{i=1}^{n}\left[ \frac{1}{m}\sum_{j=1}^{m} ||\tilde{x}_{s_{i}}^{j}(\boldsymbol{\omega}^\phi) - x_{s_{i}}^{j}||_{2}^\beta  -\frac{\lambda}{2} \frac{1}{m(m-1)}\sum_{j\neq k} ||\tilde{x}_{s_{i}}^{j}(\boldsymbol{\omega}^\phi) -\tilde{x}_{s_{i}}^{k}(\boldsymbol{\omega}^\phi)||_{2}^\beta \right]$
        \State (\textbf{Reward loss}) Compute reward loss
         \State $\hat{\mathcal{L}}_{R}(\phi) = \frac{1}{nm}\sum_{i=1}^{n} \sum_{j=1}^m R(\hat{x}_{\theta}(\tilde{x}^j_{t_i}, c_i; \omega_i), c_i)$
        \State (\textbf{Total loss}) Compute total loss
        \State $\tilde{\mathcal{L}}_{tot}(\phi) = \hat{\mathcal{L}}_{\beta,\lambda}(\phi) + \gamma_R \hat{\mathcal{L}}_{R}(\phi)$
        \State Update $\phi \leftarrow \phi - \eta \nabla_{\phi} \tilde{\mathcal{L}}_{tot}(\phi$
    \Until{convergence}
    \State \textbf{Output:} Optimized guidance network parameters $\phi$
\end{algorithmic}
\label{alg:learning_to_guide_with_rewards}
\end{algorithm}

\section{Experimental details}
\label{app_sec:experimental_details}

\paragraph{Noise process.} For all the experiments, we use recitifed flow noise process~(\eqref{eq:forward}) where $\alpha_t = (1-t)$ and $\sigma_t=t$.

\paragraph{Guidance network architecture for MoG, ImageNet and CelebA.} For guidance network $\omega(s,t,c;\phi) = \omega^\phi_{(c,(s,t))}$ we use a $6$ layers MLP with hidden size of $64$ followed by an output layer transforming it to a dimension $1$. As an activation we use "GeLU". We also apply a "ReLU" activation after the last layer of the guidance network to ensure that the gudiance weights are non-negative. For 2D experiments, we do not use ReLU activation and allow the weights to be negative.

The time which is processed by the weight network is first converted to $\log \text{SNR}$, i.e. $s \rightarrow \log \text{SNR}(s)$ and $t \rightarrow \log \text{SNR}(t)$. We do not use sinusoidal embedding as typically is done with diffusion models, since we found that it worked significantly worse. Then the two time-steps are concatenated and passed through to a 2-layer MLP with hidden dimension 256 and an output dimension 512 and GeLU activation. We also use a dropout with rate $0.3$ in the middle of this MLP.

\paragraph{2D MoG.} We use $4$ layer MLP with hidden dimension $64$ and GeLU activation for the denoiser network. The time $t$ for the backbone, is first embedded into $\log \text{SNR}$ and then we use sinusoidal embedding dimension 128. We use adam optimizer with learning rate $10^{-4}$ and batch size $128$. We use maximum norm clipping for Adam to be $1$. When we do pretraining in the \emph{well-trained} regime, we use $10k$ iterations, and when we do pretraining in the \emph{under-trained} regime, we use $250$ iterations. We train guidance network for 1000 iterations, with learning rate $5e-4$, norm clipping $1$ and Adam optimizer. We use $m=32$ particles. We use batch size $128$. We use backwards DDIM~(\eqref{eq_app:mean_sigma}) with $\varepsilon = 1$ in order to sample $x_s$ with guidance~(\eqref{eq:guided_forward_sample_2}). To produce \cref{fig:2d_results} and \cref{fig:2d_guided_samples}, we sample data with velocity sampler with $10$ steps. We use $4096$ samples in total.

\paragraph{ImageNet.} We pretrain the U-Net model~\citep{ronneberger2015u} similar to DDPM~\citep{ho2020denoising}  which has channel dimension 192, 3 residual blocks, channel multiplier (1, 2, 3, 4), attention which is applied with (False, False, True, False) residual blocks (whenever it is True, we use attention), dropout which is applied at (False, True, True, True) residual blocks (whenever it is True, we use dropout). The dropout rate is $0.1$. We also use sigmoid weighting~\citep{kingma2023understanding} with bias $b=2$. The model predicts the velocity and the loss is based on velocity. We train it with batch size $128$ for $7M$ iterations. We use Adam optimizer with learning rate $1e-5$ and norm clipping $1$. We use EMA decay of $0.9999$.

For training guidance we reload the pretrained diffusion model with EMA parameters and freeze it. We train the guidance network for $100K$ iterations with batch size $B=256$. We use $m=4$ number of particles and $\beta=1.75,\lambda=1$ parameters. We use learning rate $1e-5$ with adam optimizer and norm clipping $1$. We use EMA decay of $0.9999$. For self-consistency loss~(\eqref{eq:distributional_objective}), in order to produce $\tilde{x}_s(\boldsymbol{\omega)}$ in~(\eqref{eq:guided_forward_sample_2}), we use DDIM sampler~(\eqref{eq_app:mean_sigma}) with $\varepsilon=1$. We sweep over $\delta \in [0.01,0.1,0.2,0.3]$, $S_{\min} \in [0.01, 0.1,0.2,0.3]$ parameters. We use $\zeta=1e-2$ which is the same parameter as the safety parameter for the original diffusion model. This safety parameter in the original diffusion model just controls the time interval to be $(\zeta, 1-\zeta)$ instead of $(0,1)$, for numerical stability. To sample data, we use DDIM~(\eqref{eq:guided_forward_sample_2}) with $\varepsilon=0$ for all the experiments. During training, we track FID on a small subset of $2048$ images. We observed that the performance over time could be unstable for some hyperparameters, therefore we use this training-time FID to keep the best checkpoint for every hyperparameter. We report the performance for the best hyperparameters based on FID, we use $50k$ samples for FID computation. On top of that, we also train the simplified $\ell_2$ method using~\Cref{app_alg:l2_algorithm} using the same methodology, except that in order to produce $\tilde{x}_s(\boldsymbol{\omega)}$ in~(\eqref{eq:guided_forward_sample_2}), we use DDIM sampler~(\eqref{eq_app:mean_sigma}) but with $\varepsilon=0$, since it led to a better empirical performance.

\paragraph{CelebA.} We pretrain the U-Net model~\citep{ronneberger2015u} similar to DDPM~\citep{ho2020denoising}  which has channel dimension 256, 2 residual blocks, channel multiplier (1, 2, 2, 2), attention which is applied with (False, True, False, False) residual blocks (whenever it is True, we use attention), dropout which is applied at (False, True, True, True) residual blocks (whenever it is True, we use dropout). The dropout rate is $0.2$. We also use sigmoid weighting~\citep{kingma2023understanding} with bias $b=1$. We do not transform time to $\log \text{SNR}$ for time embedding in the diffusion network. The model predicts the $x_0$ and the loss is based on velocity. We train it with batch size $128$ for $300K$ iterations. We use Adam optimizer with learning rate $1e-4$ and norm clipping $1$. We use EMA decay of $0.9999$.

For training guidance we reload the pretrained diffusion model with EMA parameters and freeze it. We train the guidance network for $100K$ iterations with batch size $B=256$. We use $m=4$ number of particles and $\beta=1.75,\lambda=1$ parameters. We use learning rate $1e-5$ with adam optimizer and norm clipping $1$. We use EMA decay of $0.9999$. For self-consistency loss~(\eqref{eq:distributional_objective}), in order to produce $\tilde{x}_s(\boldsymbol{\omega)}$ in~(\eqref{eq:guided_forward_sample_2}), we use DDIM sampler~(\eqref{eq_app:mean_sigma}) with $\varepsilon=1$. We sweep over $\delta \in [0.01,0.1,0.2,0.3]$, $S_{\min} \in [0.01, 0.1,0.2,0.3]$ parameters. We use $\zeta=1e-2$ which is the same parameter as the safety parameter for the original diffusion model. This safety parameter in the original diffusion model just controls the time interval to be $(\zeta, 1-\zeta)$ instead of $(0,1)$, for numerical stability. To sample data, we use DDIM with $\varepsilon=0$ for all the experiments. During training, we track FID on a small subset of $2048$ images. We observed that the performance over time could be unstable for some hyperparameters, therefore we use this training-time FID to keep the best checkpoint for every hyperparameter. We report the performance for the best hyperparameters based on FID, we use $50k$ samples for FID computation. On top of that, we also train the simplified $\ell_2$ method using~\Cref{app_alg:l2_algorithm} using the same methodology, except that in order to produce $\tilde{x}_s(\boldsymbol{\omega)}$ in~(\eqref{eq:guided_forward_sample_2}), we use DDIM sampler~(\eqref{eq_app:mean_sigma}) but with $\varepsilon=0$, since it led to a better empirical performance.

\paragraph{Guidance network architecture for MS-COCO.} For guidance network $\omega(s,t,c_{\text{CLIP}}, c_{\text{T5}};\phi)=\omega^\phi_{c_{\text{CLIP}}, c_{\text{T5}, (s,t)}}$ we use a $6$ layers MLP with hidden size of $512$ followed by an output layer transforming it to a dimension $1$. As an activation we use "GeLU". We also apply a "ReLU" activation after the last layer of the guidance network to ensure that the guidance weights are non-negative.

The time steps $t$ and $s$ are each processed by separate networks. We do not use sinusoidal embedding to get the time embeddings. We also do not convert time to $\log \text{SNR}$ as we did for ImageNet and CelebA. Each of the two time-steps are passed through to a 2-layer MLP with hidden dimension and output dimension of $256$ and a SiLU activation. The two timestep embeddings are then concatenated. We also pass  CLIP and T5 embeddings through a similar MLP with an output dimension of $256$ and $512$ respectively. Both time embedding and CLIP embedding are concatenated to a dimension $512$ and then concatenated with the text embedding. This is passed through the MLP layers described above.

\paragraph{MS-COCO.} We pretrain a 1.05B parameter Flux text-to-image backbone~\citep{labs2025flux}, which uses an architecture based on~\citet{esser2024scaling} and consists of 30 Diffusion Transformer (DiT) blocks~\citep{peebles2023scalable} and 15 MMDiT blocks~\citep{esser2024scaling}.
 We do not use any weighting, nor do we transform time with logSNR. The model is trained with the standard flow matching loss and predicts velocity.  The CLIP embeddings used for training the model are extracted from the standard ViT-L/14 CLIP transformer architecture. The T5 embeddings are extracted from the standard T5-XXL architecture.

 We load the pretrained flow matching model with EMA parameters and freeze it to train the guidance network. We train the guidance network for $50K$ iterations with a batch size of $256$. 
 We set the number of particles as $m=4$, $\beta = 1.75$ and $\lambda = 1.0$. We use learning rate of $8e-5$ with Adam optimizer. We set the Adam parameters $\beta=(0.9, 0.999)$, $\epsilon=1e-8$, and weight decay=$0.01$.
 For self-consistency loss~(\eqref{eq:distributional_objective}), in order to produce $\tilde{x}_s(\boldsymbol{\omega)}$ in~(\eqref{eq:guided_forward_sample_2}), we use DDIM sampler~(\eqref{eq_app:mean_sigma}) with $\varepsilon=0$.
 We set $\delta=0.2$,  $S_{\text{min}}=0.01$ and safety parameter $\zeta=1e-2$ while training the model.
To sample data from flow matching model, we use the standard Euler ODE solve in the time interval $[0, 1]$ and use $128$ sampling steps. To select the best checkpoint, we tracked CLIP Score over $3K$ samples and then used the checkpoint corresponding to the best CLIP Score on this subset to compute CLIP Score over $10K$ samples.

\section{Additional experiments and results}
\label{app_sec:additional_experiments}

\subsection{2D Mixture of Gaussians (MoG)}
\label{sec_exp:2d_results}

\begin{figure}
    \centering
        \includegraphics[width=.9\linewidth]{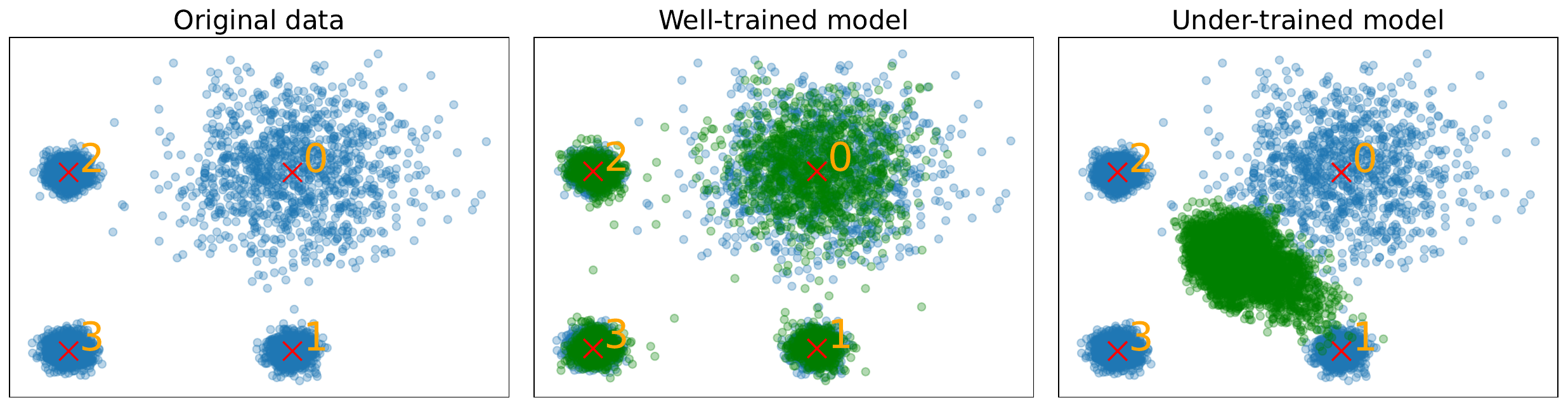}
    \caption{\textbf{Mixture of Gaussians data}. Red cross corresponds to the locations of the mean, while the numbers denote respective numbering of mixture components. The blue color corresponds to the data, while green color corresponds to the samples from diffusion model.}
    \label{fig:2d_moe_data}
\end{figure}

We consider a 4 components MoG with the means $(10, 10)$, $(-10, 10)$, $(10, -10)$ and $(-10, -10)$, as well as the diagonal covariances $\sigma^2 \Id$ where the corresponding variances $\sigma^2$ are equal to $5,1,1,1$. The data is visualized in Figure~\ref{fig:2d_moe_data}, left.

We study two regimes, \emph{well-trained} and \emph{under-trained}. In a \emph{well-trained} regime, the diffusion model is trained for long enough in order to produce the samples close to the training distribution $p_{0|c}$, while in \emph{under-trained}, it is trained for a small amount of iterations and produces samples far away from the training distribution. See~\Cref{fig:2d_moe_data}, center and right for corresponding samples. Please refer to Appendix~\ref{app_sec:experimental_details} for all experimental details.

\begin{figure}
    \centering

    % Big A and B above
    \makebox[0.05\linewidth][c]{\LARGE \textbf{A}}%
    \hfill
    \makebox[0.95\linewidth][c]{\LARGE \textbf{B}}%
   
    \begin{subfigure}{0.45\linewidth}
        \centering
        \includegraphics[width=.95\linewidth]{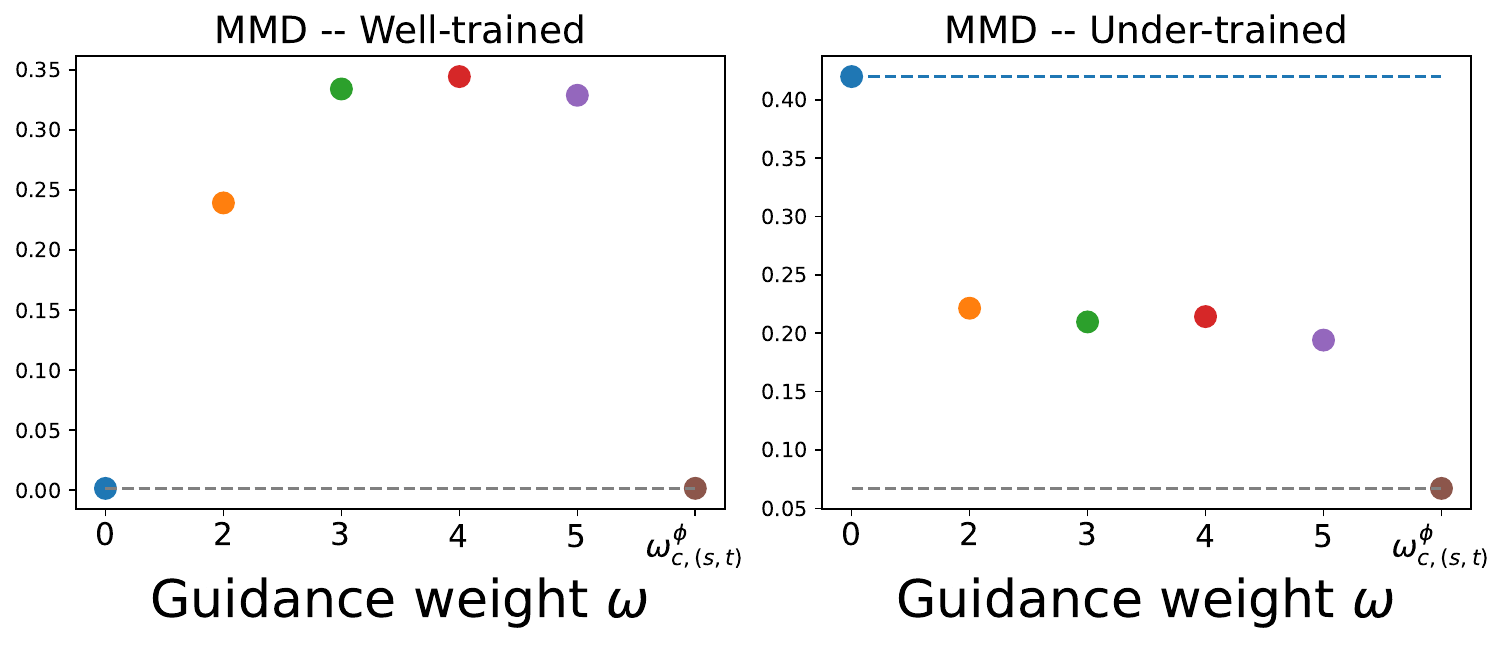}
        \label{fig:2d_mmds}
    \end{subfigure}
    \hfill
    \vrule width 1pt % vertical line (adjust width as needed)
    \hfill
    \begin{subfigure}{0.45\linewidth}
        \centering
        \includegraphics[width=.95\linewidth]{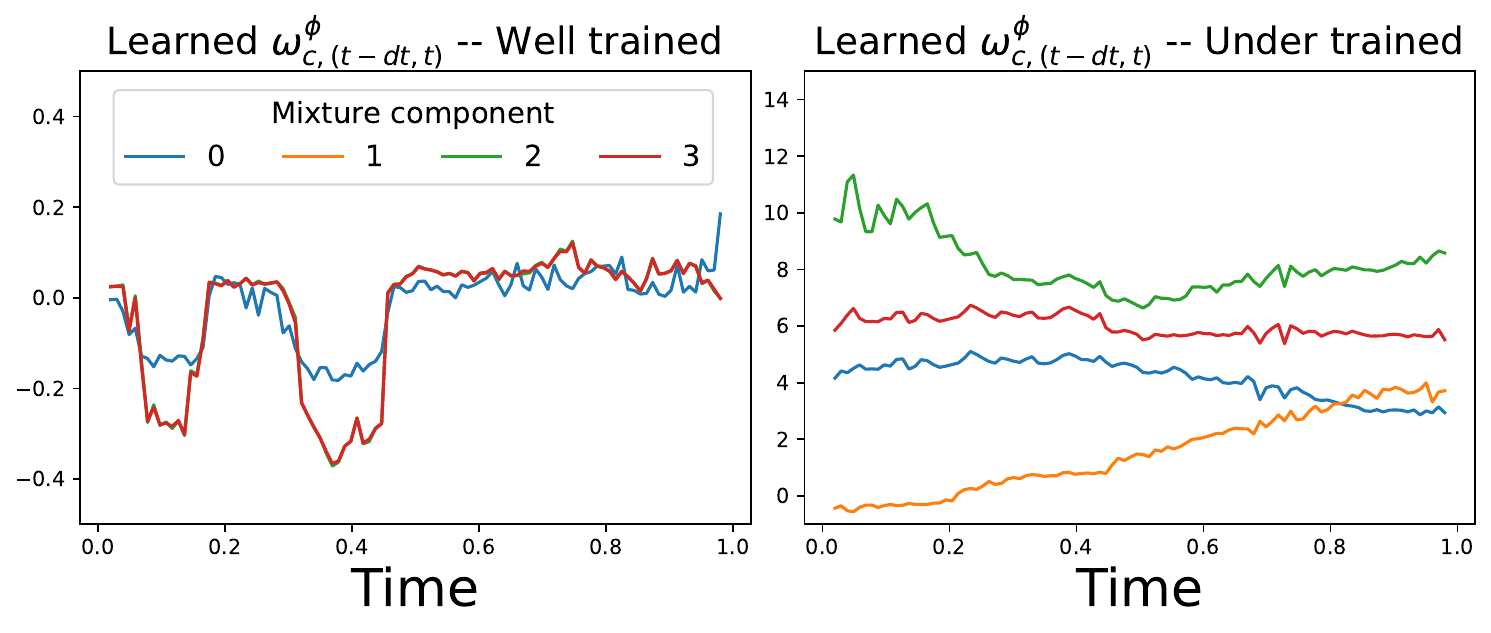}
        \label{fig:2d_learned_w}
    \end{subfigure}
    \caption{\textbf{A}: $\MMD$ between samples ($4096$) from data distribution and samples produced by a diffusion model with different guidance weight. Blue dashed line denotes the performance of unguided model, while brown dashed line denotes the performance of a model guided with learned $\omega^\phi_{c,(t-dt,t)}$. \textbf{B}: learned guidance weight  $\omega^\phi_{c,(t-dt,t)}$ from~\Cref{alg:learning_to_guide} for different mixture components.}
    \label{fig:2d_results}
\end{figure}

\begin{figure}
    \centering
        \includegraphics[width=.7\linewidth]{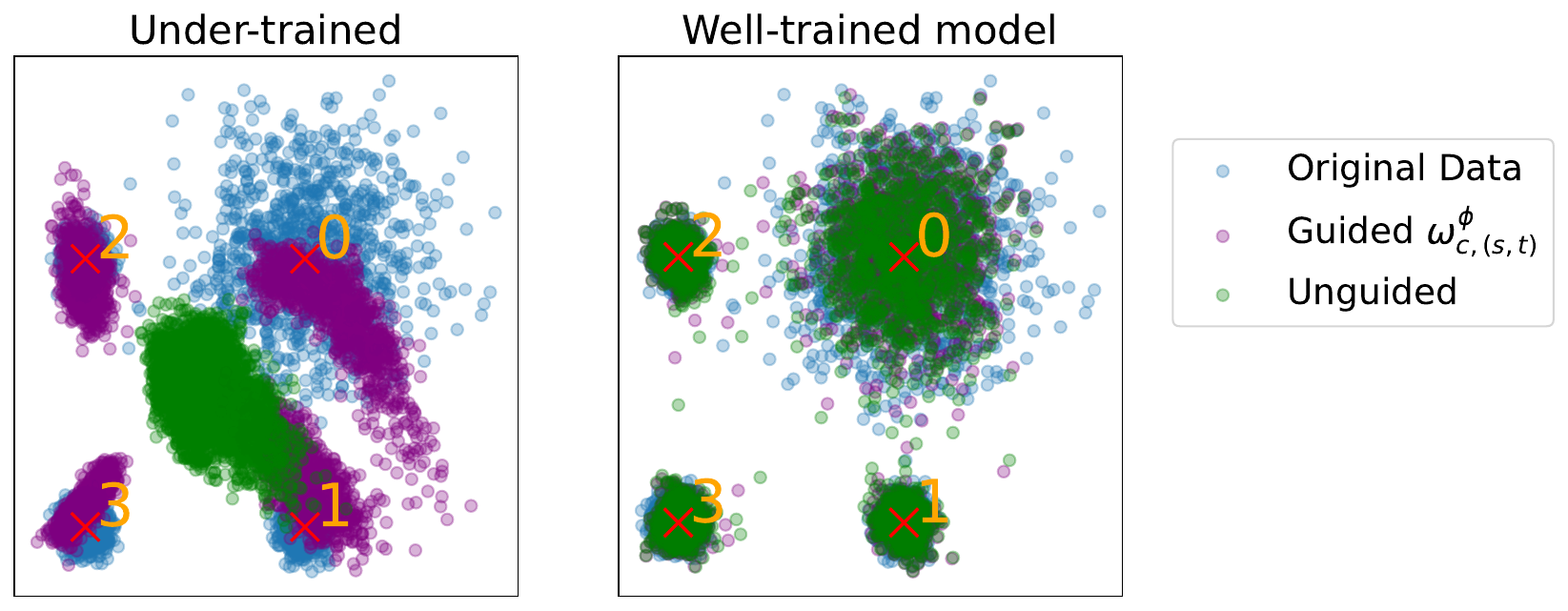}
    \caption{\textbf{Samples with and without guidance}. In blue, we show the original data, in green we show samples from unguided model while in purple we show samples from the guided model with learned guidance $\omega^\phi_{c,(s,t)}$ with~\Cref{alg:learning_to_guide}. First column corresponds to \emph{under-trained} regime while second column corresponds to \emph{well-trained} one.}
    \label{fig:2d_guided_samples}
\end{figure}

We train $\omega^\phi_{c,(s,t)}$ with~\Cref{alg:learning_to_guide}. The results are provided in~\Cref{fig:2d_results}. In \Cref{fig:2d_results}, A, we visualize the $\MMD$ between samples from data distributions and samples produced by diffusion model with different constant guidance weights $\omega$. In the \emph{well-trained} regime, as we increase the guidance weight $\omega$, it degrades the $\MMD$, because it pushes the samples away from the distribution. What is intriguing is that in \emph{under-trained} regime, $\MMD$ decreases as the guidance weight $\omega$ increases, achieving the lowest value with the learned guidance weight $\omega^\phi_{c,(s,t)}$. This finding highlights one of our main message -- CFG can be used to correct the mismatch of the sampled distribution and the data distribution. In \Cref{fig:2d_results}, B, we visualize the learned  $\omega^\phi_{c,(t-dt,t)}$, where $dt$ is a time discretization. In the \emph{well-trained} regime, the learned guidance weights tend to be close to $0$, while in \emph{under-trained} regime, the learned guidance weights reach quite high values. In~\Cref{fig:2d_guided_samples}, we visualize the corresponding samples. This finding highlights that in an under trained regime, high guidance weights are required.

\end{document}